\documentclass{article}

\usepackage[nonatbib,final]{neurips_2022}

\usepackage[utf8]{inputenc} 
\usepackage[T1]{fontenc}    
\usepackage{hyperref}       
\hypersetup{
colorlinks= true,
urlcolor = blue,
linkcolor = red,
citecolor = blue,
}
\usepackage{url}            
\usepackage{booktabs}       
\usepackage{amsfonts}       
\usepackage{nicefrac}       
\usepackage{microtype}      
\usepackage{xcolor}         
\usepackage{xspace}

\usepackage{graphicx}
\usepackage{caption}
\usepackage{threeparttable}
\usepackage{subcaption}
\usepackage{bm}
\usepackage{amstext}
\usepackage{amsmath}
\usepackage{amsthm}
\usepackage{amssymb}
\usepackage{pifont}
\usepackage{enumitem}
\usepackage{pifont}
\usepackage{color}
\usepackage[ruled,linesnumbered]{algorithm2e}
\usepackage{wrapfig}

\newcommand{\bw}{\mathbf{w}}
\newcommand{\bx}{\mathbf{x}}
\newcommand{\bz}{\mathbf{z}}

\newcommand{\by}{\mathbf{y}}
\newcommand{\be}{\mathbf{e}}
\newcommand{\bg}{\mathbf{g}}
\newcommand{\bv}{\mathbf{v}}

\newcommand{\1}{\mathbf{1}}

\newcommand{\bW}{{\rm \mathbf{W}}}
\newcommand{\bX}{{\rm \mathbf{X}}}

\newcommand{\bE}{{\mathbf{E}}}
\newcommand{\bZ}{{\mathbf{Z}}}

\newcommand{\method}{\texttt{TransTab}\xspace}
\newcommand{\clmethod}{\texttt{VPCL}\xspace}

\title{\method: Learning Transferable Tabular Transformers Across Tables}

\author{Zifeng Wang \\
  UIUC \\
  \texttt{zifengw2@illinois.edu} \\
  \And
  Jimeng Sun \\
  UIUC \\
  \texttt{jimeng@illinois.edu} \\
}

\begin{document}

\maketitle
\begin{abstract}
Tabular data (or tables) are the most widely used data format in machine learning (ML). However, ML models often assume the table structure keeps fixed in training and testing. Before ML modeling, heavy data cleaning is required to merge disparate tables with different columns. This preprocessing often incurs significant data waste (e.g., removing unmatched columns and samples). How to learn ML models from multiple tables with partially overlapping columns? How to incrementally update ML models as more columns become available over time? Can we leverage model pretraining on multiple distinct tables? How to train an ML model which can predict on an unseen table? 

To answer all those questions, we propose to relax fixed table structures by introducing a Transferable Tabular Transformer (\method) for tables. The goal of \method is to convert each sample (a row in the table) to a generalizable embedding vector, and then apply stacked transformers for feature encoding. One methodology insight is combining column description and table cells as the raw input to a gated transformer model. The other insight is to introduce supervised and self-supervised pretraining to improve model performance. We compare \method with multiple baseline methods on diverse benchmark datasets and five oncology clinical trial datasets. Overall, \method ranks 1.00, 1.00, 1.78 out of 12 methods in supervised learning, feature incremental learning, and transfer learning scenarios, respectively; and the proposed pretraining leads to 2.3\% AUC lift on average over the supervised learning.

\end{abstract}


\section{Introduction}\label{sec:introduction}
Tabular data are ubiquitous in healthcare, engineering, advertising, and  finance~\cite{wang2017deep,yoon2020vime,zhang2020customer,wang2020data}. 
They are often stored in a relational database as  tables or spreadsheets. Table rows represent the data samples, and columns represent the feature variables of diverse data types (e.g., categorical, numerical, binary, and textual).  
Recent works enhance tabular ML modeling using deep networks \cite{huang2020tabtransformer,padhi2021tabular,cholakov2022gatedtabtransformer,wang2021survtrace} or designing self-supervision \cite{yoon2020vime, ucar2021subtab,somepalli2021saint,bahri2022scarf}. Those existing works require the same table structure in training and testing data. However, there can be multiple tables sharing partially overlapped columns in the real world. Hence, learning across tables is inapplicable. The traditional remedy is to perform data cleaning by removing non-overlapping columns and mismatched samples before training any ML models, which waste data resources~\cite{nargesian2018table,zhu2019josie,dong2021efficient}. Therefore, learning across tables with disparate columns and transferring knowledge across tables are crucial to extending the success of deep learning/pretraining to the tabular domain.

Tables are highly structured yet flexible. The first step to achieve learning across tables is to rethink the \emph{basic elements} in tabular data modeling. In computer vision, the basic elements are pixels \cite{he2021masked} or patches, \cite{dosovitskiy2020image,bao2021beit}; in natural language processing (NLP), the basic elements are words \cite{mikolov2013efficient} or tokens \cite{schuster2012japanese,devlin2018bert}. In the tabular domain, it is natural to treat cells in each column as independent elements. Columns are mapped to unique indexes then models take the cell values for training and inference. The premise of this modeling formulation is to keep the same column structure in all the tables. But tables often have divergent protocols where the nomenclatures of columns and cells differ. By contrast, our proposed work contextualizes the columns and cells. For example, previous methods represent a cell valued \textit{man} under the column \textit{gender} by $0$ referring to the codebook $\{\text{man}:0,\text{woman}:1\}$.  Our model converts the tabular input into a sequence input (e.g., \textit{gender is man}),  which can be modeled with downstream sequence models. We argue such featurizing protocol is generalizable across tables, thus enabling models to apply to different tables.

\begin{figure}[t]
\centering
\includegraphics[width=1\textwidth]{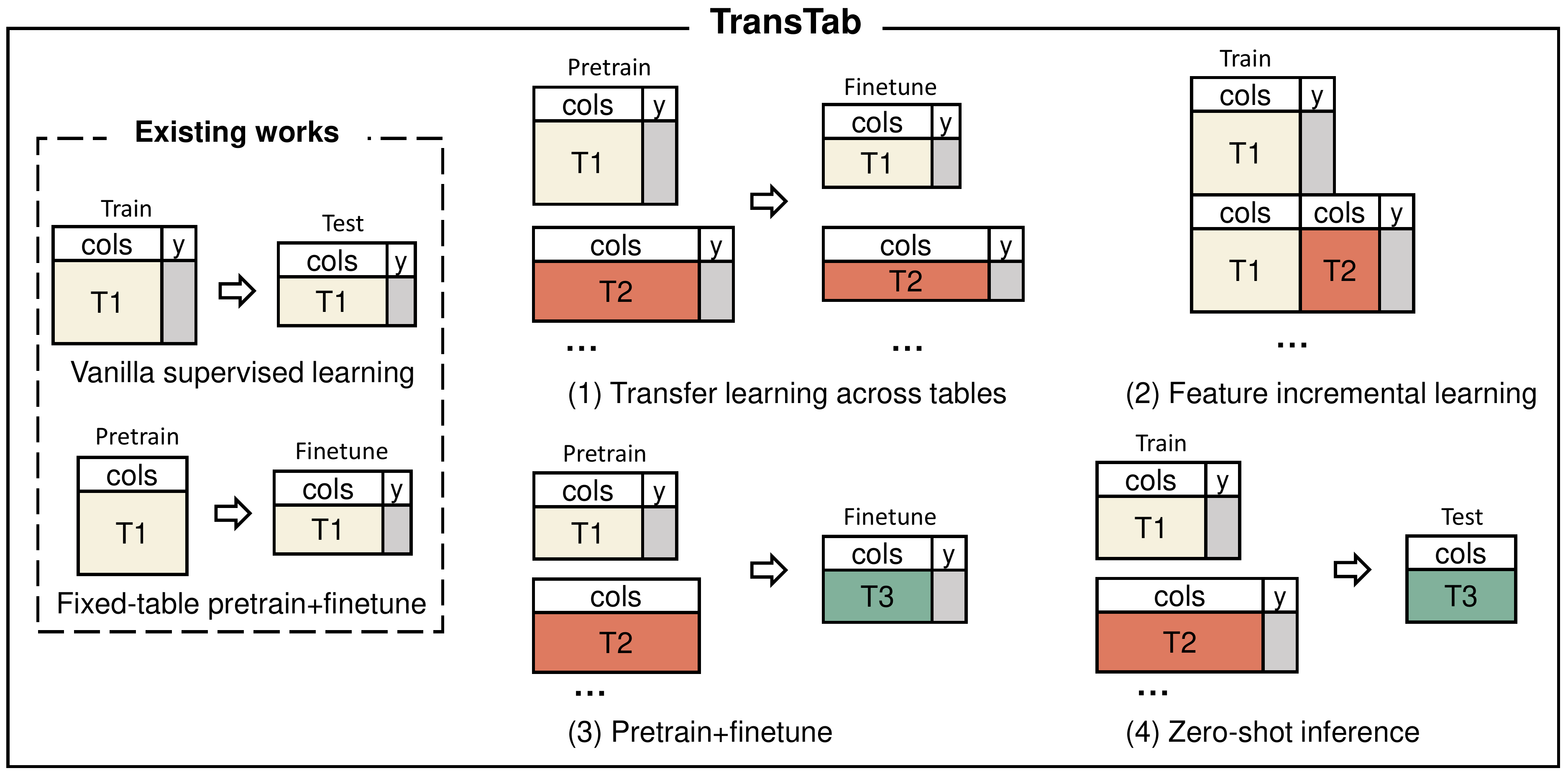}
\caption{The demonstration of ML modeling on different tabular data settings. Previous tabular methods only do vanilla supervised training or pretraining on the same table due to they only accept \textbf{fixed-column tables}. By contrast, \method covers more new tasks (1) to (4) as it accepts \textbf{variable-column} tables. Details are presented in \S \ref{sec:inference}. \label{fig:1}}
\vspace{-1em}
\end{figure}

In a nutshell, we propose \textbf{Trans}ferable \textbf{Trans}formers for \textbf{Tab}ular analysis (\method), a versatile tabular learning framework \footnote{Our package is available at \url{https://github.com/RyanWangZf/transtab} with documentation at \url{https://transtab.readthedocs.io/en/latest/}.}. \method applies to multiple use cases as shown in Fig. \ref{fig:1}. 
The key contributions behind \method are 
\begin{itemize}[leftmargin=*, itemsep=0pt, labelsep=5pt]
\item A systematic featurizing pipeline considering both column and cell semantics which is shared as the fundamental protocol across tables.
\item \textbf{V}ertical-\textbf{P}artition \textbf{C}ontrastive \textbf{L}earning (\clmethod) that enables pretraining on multiple tables and also allows finetuning on target datasets.
\end{itemize}
As shown by Fig. \ref{fig:1}, due to the fixed-column assumption, all existing works only handle supervised learning or pretraining on the same-structure tables. On the contrary, \method relaxes this assumption and applies to four additional scenarios, which we will elaborate on in \S \ref{sec:inference}.

\section{Method}
\begin{figure}[t]
\centering
\includegraphics[width=1.0\textwidth]{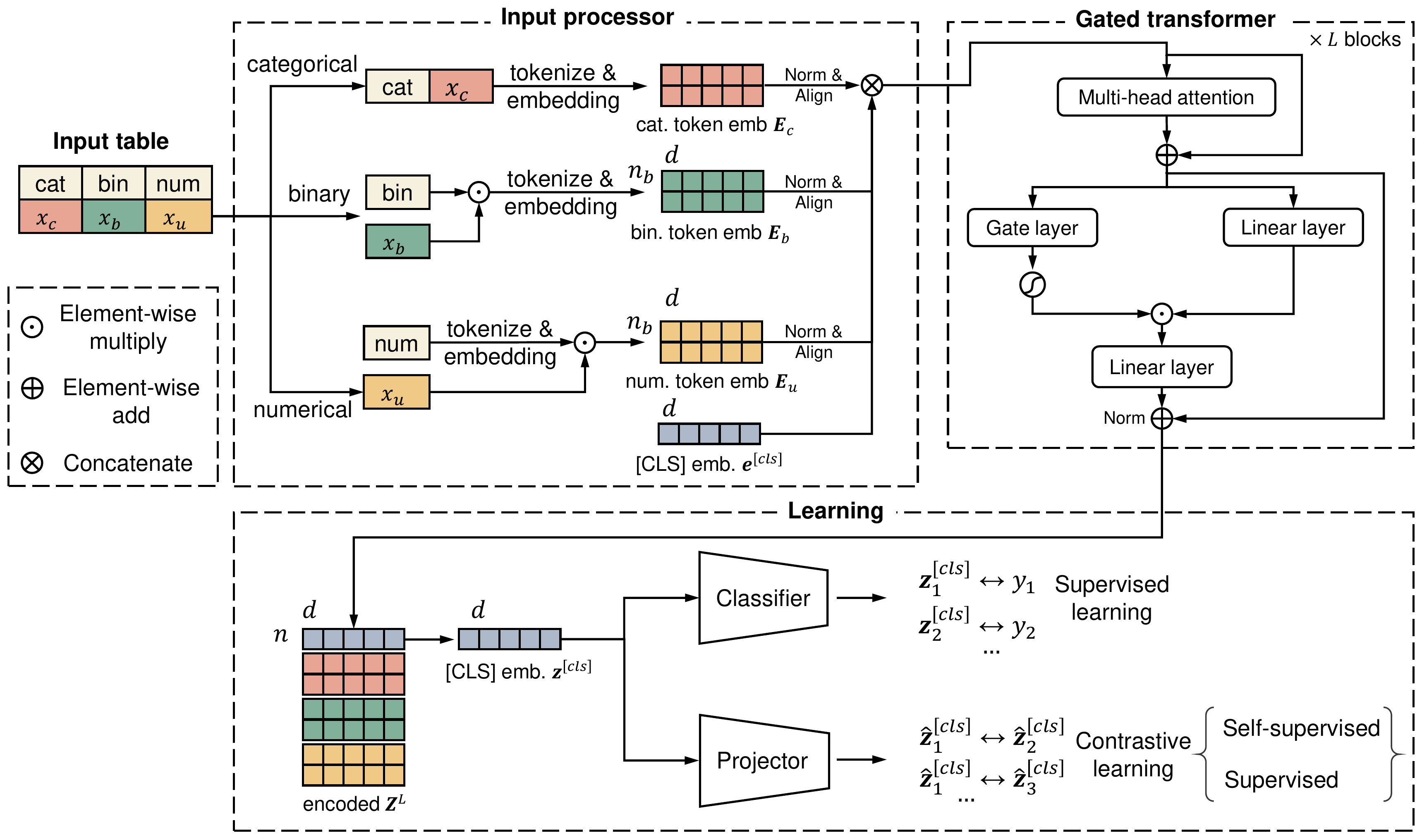}
\caption{The demonstration of \method framework. the \textit{input processor} encodes the sample into the token-level embedding $\bE$; the \texttt{[cls]} embedding $\bz^{[cls]}$ in the representation $\bZ^L$ after $L$ \textit{gated transformer} layers is used for prediction and learning. In supervised learning, $\bz^{[cls]}$ is leveraged by a classifier to make predictions of target $y$; in contrastive learning, the projected $\hat{\bz}^{[cls]}$ is is used for self or supervised contrastive loss. \label{fig:2}}
\vspace{-1em}
\end{figure}

In this section, we present the details of \method. Fig. \ref{fig:2} illustrates its workflow including the following key components: 1) The \textit{input processor} featurizes and embeds arbitrary tabular inputs to token-level embeddings; 2) The stacked gated transformer layers further encode the token-level embeddings; 3) Finally, the {\it learning} module includes a  {\it classifier} trained on labeled data and a {\it projector} for contrastive learning. Next we will present the details of each component.

\subsection{Application scenarios of \method} \label{sec:inference}
Before presenting our method in details, we first introduce four novel applications scenarios which are tractable by \method, as shown in Fig. \ref{fig:1}. Suppose we aim to predict the treatment efficacy for breast cancer trials using multiple clinical trial tables, here are several scenarios we often encounter.

\textbf{S(1) Transfer learning.} We collect data tables from multiple cancer trials for testing the efficacy of the same drug on different patients. These tables were designed independently with overlapping columns. How do we learn ML models for one trial by leveraging tables from all trials?

\textbf{S(2) Incremental learning.} Additional columns might be added over time. For example, additional features are collected across different trial phases. How do we update the ML models using tables from all trial phases?

\textbf{S(3) Pretraining+Finetuning.} The trial outcome label (e.g., mortality) might not be always available from all table sources. Can we benefit pretraining on those tables without labels? How do we finetune the model on the target table with labels?


\textbf{S(4) Zero-shot inference.} We model the drug efficacy based on our trial records. The next step is to conduct inference with the model to find patients that can benefit from the drug. However, patient tables do not share the same columns as  trial tables so direct inference is not possible.

Overall, we witness that the assumption of fixed table structure is the obstacle to use ML for various applications. Next we will present \method and demonstrate how it addresses these scenarios.

\subsection{Input processor for columns and cells} \label{sec:processor}
We build the input processor (1) to accept variable-column tables (2) to retain knowledge across tabular datasets. 
The idea is to convert tabular data (cells in columns) into a sequence of semantically encoded tokens. We utilize the following observation to create the sequence: the column description (e.g., column name) decides the meaning of cells in that column. For example, if a cell in  column \textit{smoking history} has value $1$, it indicates the individual has a smoking history. Similarly, cell value $60$ in column  \textit{weight} indicates 60 kg in weight instead of 60 years old. Motivated by the discussion, we propose to include column names into the tabular modeling. As a result, \method treats any tabular data as the composition of three elements: text (for categorical \& textual cells and column names),  continuous values (for numerical cells), and boolean values (for binary cells) . Fig. \ref{fig:2} illustrates a visual example of how these elements are leveraged to process the four basic types of features: categorical/textual \texttt{cat}, binary \texttt{bin}, and numerical \texttt{num}. 

\textbf{Categorical/Textual feature.} A category or textual feature contains a sequence of text tokens. For the categorical feature \texttt{cat}, we concatenate the column name with the feature value $x_c$, which forms as a sequence of tokens. This sentence is then tokenized and matched to the token embedding matrix to generate the feature embedding $\bE_c \in \mathbb{R}^{n_c \times d}$ where $d$ is the embedding dimension and $n_c$ is the number of tokens.

\textbf{Binary feature.} The binary feature \texttt{bin} is usually an assertive description and its value $x_b \in \{0,1\}$. If $x_b=1$, then \texttt{bin} is tokenized and encoded to the embeddings $\bE_{b}\in\mathbb{R}^{n_b\times d}$; if not, it will not be processed to the subsequent steps. This design significantly reduces the computational and memory cost when the inputs have high-dimensional and sparse one-hot features.

\textbf{Numerical feature.} We do not concatenate column names and values for numerical feature because the tokenization-embedding paradigm was notoriously known to be bad at discriminating numbers \cite{lin2020birds}. Instead, we process them separately. \texttt{num} is encoded as same as \texttt{cat} and \texttt{bin} to get $\bE_{u,col} \in \mathbb{R}^{n_u\times d}$. We then multiply the numerical features with the column embedding to yield the numerical embedding as $\bE_{u} = x_u \times \bE_{u,col}$\footnote{$x_u$ is standardized or normalized in preprocessing.}, which we identify gets an edge on more complicated numerical embedding techniques empirically.


At last, $\bE_c$, $\bE_u$, $\bE_b$ all pass the layer normalization \cite{ba2016layer} and the same linear layer to be aligned to the same space, then are concatenated with \texttt{[cls]} embedding to yield $\bE = \tilde{\bE}_c \otimes \tilde{\bE}_u \otimes \tilde{\bE}_b \otimes \be^{[cls]}$. 


As a result, all cell values are contextualized regarding the corresponding column properties thus the semantic meaning of one element can vary depending on the context composition. This formulation benefits the knowledge transfer across tables a lot. For example, \textit{previously smoked} depicts the same thing as \textit{smoking history}. Previous methods never capture this connection while it is possible for \method to learn to recognize that $1$ under both columns are equivalent.

\subsection{Gated transformers}
The gated tabular transformer is an adaption of the classical transformer in NLP \cite{vaswani2017attention}. It consists of two main components: multi-head self-attention layer and gated feedforward layers. The input representation $\bZ^l$ at the $l$-th layer is first adopted for exploring interactions between features:
\begin{align}
   \bZ^l_{\text{att}} = \texttt{MultiHeadAttn}(\bZ^l) = [\texttt{head}_1,\texttt{head}_2,\dots,\texttt{head}_h]\bW^O, \\
    \texttt{head}_i = \texttt{Attention}(\bZ^l\bW^Q_i, \bZ^l\bW^K_i, \bZ^l\bW^V_i), \label{eq:attention_head}
\end{align}
where $\bZ^0 = \bE$ at the first layer; $\bW^O \in \mathbb{R}^{d\times d}$; $\{\bW^Q_i,\bW^K_i,\bW^V_i\}$ are weight matrices (in $\mathbb{R}^{d\times \frac{d}{h}}$) of query, key, value of the $i$-th head self-attention module.


The multi-head attention output $\bZ^l_{\text{att}}$ is further transformed by a token-wise gating layer as $\bg^l = \sigma(\bZ_{\text{att}}^l \bw^G)$, where $\sigma(\cdot)$ is a sigmoid function; $\bg^l \in [0,1]^n$ controls the magnitude of each token embedding before $\bZ_{\text{att}}$ goes to the linear projection. This gates then filters the linear layer output
\begin{equation}
    \bZ^{l+1} = \texttt{Linear}\left((\bg^l \odot \bZ^l_{\text{att}})\oplus \texttt{Linear}(\bZ_{\text{att}}^l) \right)
\end{equation}
to obtain the transformer output $\bZ^{l+1} \in \mathbb{R}^{n\times d}$. This mechanism is learnt to focus on important features by redistributing the attention on tokens. The final \texttt{[cls]} embedding $\bz^{[cls]}$ at the $L$-th layer is used by the classifier for prediction.

\subsection{Self-supervised and supervised pretraining of \method}
The input processor accepts variable-column tables, which opens the door for tabular pretraining on heterogeneous tables. In detail, \method is feasible for \textit{self-supervised} and \textit{supervised pretraining}.

\begin{wrapfigure}{R}{0.5\textwidth}
\centering
\includegraphics[width=0.5\textwidth]{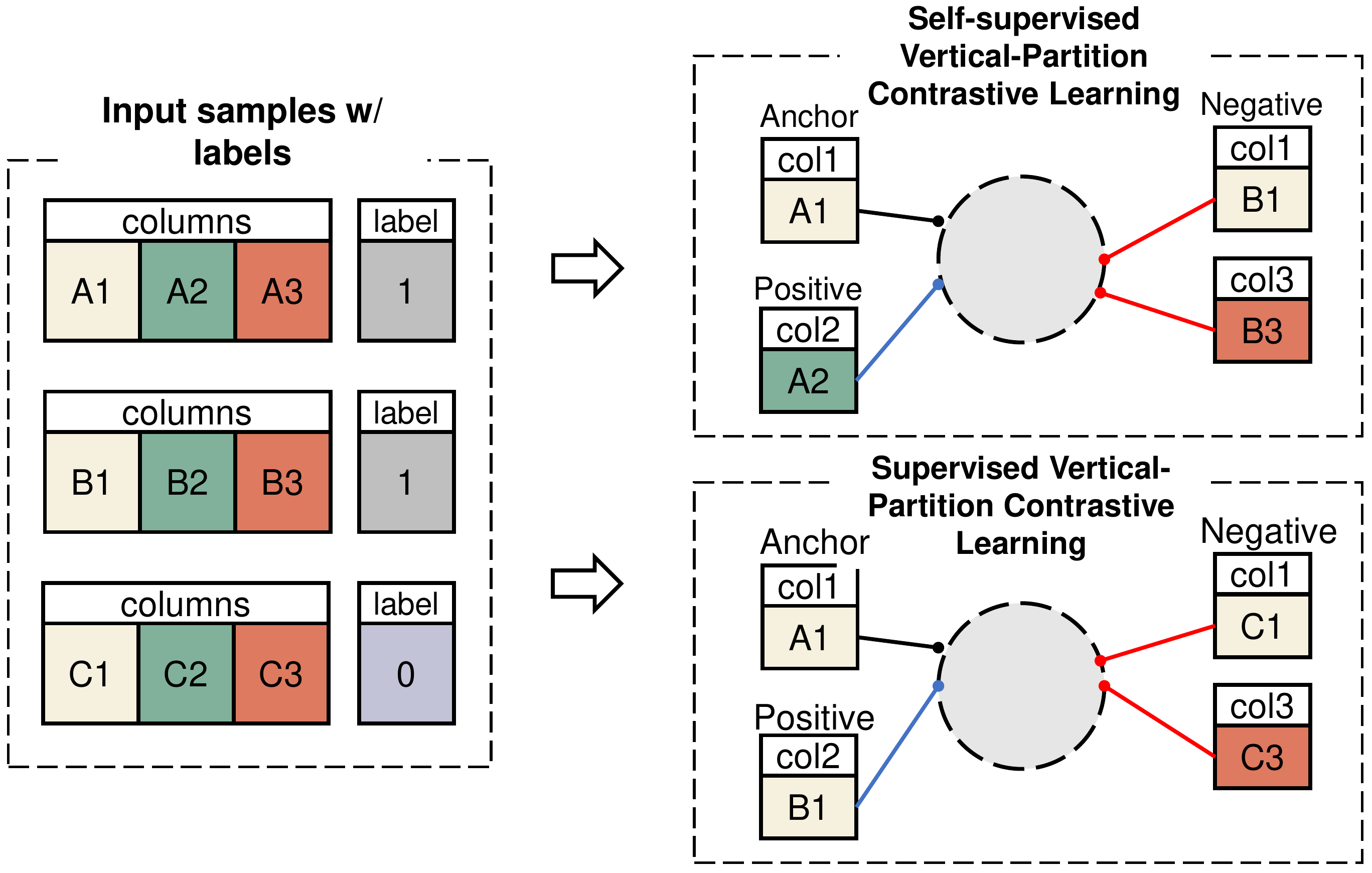}
\caption{The demonstration of contrastive learning methods (different pieces can either be distinct or be overlapped partially). Self-\clmethod: Positive pairs are partitions of the same sample; \clmethod: Positive pairs are partitions of the sample belonging to the same class. \label{fig:3}}
\vspace{-1em}
\end{wrapfigure}

\textbf{Self-supervised \clmethod.} Most SSL tabular methods work on the whole fixed set of columns \cite{yoon2020vime,darabi2021contrastive, bahri2022scarf}, which take high computational costs and are prone to overfitting. Instead, we take tabular vertical partitions to build positive and negative samples for CL under the hypothesis that the powerful representation should model view-invariant factors. In detail, we subset columns as illustrated by Fig. \ref{fig:3} where Self-\clmethod is on the top right. Suppose a sample $\bx_i = \{\bv_i^1,\dots,\bv_i^K\}$ with $K$ partitions $\bv_i^k$. Neighbouring partitions can have
overlapping regions which are justified by the percentage of columns of the partition. Self-\clmethod takes partitions from the same sample as the positive and others as the negative:

\begin{equation}\label{eq:MSSCL}
    \ell(\bX) = -\sum_{i=1}^B \sum_{k=1}^K \sum_{k^\prime \neq k}^K \log \frac{\exp\psi(\bv_i^k, \bv_i^{k^\prime})}{\sum_{j=1}^B \sum_{k^\dag=1}^K \exp \psi(\bv_i^k, \bv_j^{k^\dag})},
\end{equation}

where $B$ is the batch size; $\psi(\cdot,\cdot)$ is the cosine similarity function. $\psi$ applies to $\hat{\bz}^{[cls]}$ which is the linear projection of partition $\bv$'s embedding $\bz^{[cls]}$. Compared with vanilla CL like SCARF \cite{bahri2022scarf}, Self-\clmethod significantly expand the positive and negative sampling for learning more robust and rich embeddings. What is more, this vertical partition sampling is extremely friendly to column-oriented databases \cite{stonebraker2018c} which support the fast querying a subset of columns from giant data warehouses. For the sake of computational efficiency, when $K>2$, we randomly sample two partitions.

\textbf{Supervised \clmethod.} When we own labeled tabular data for pretraining, one natural idea would be taking task-specific predicting heads for pretraining on vanilla supervised loss, e.g., cross-entropy loss. In finetuning, these heads are dropped and a new head will be added on top of the pretrained encoder. However, we argue it is suboptimal and may undermine the model transferability. The reason behind is that tabular datasets vary dramatically in size, task definition, and class distributions. Pretraining \method using supervised loss inevitably causes the encoder biased to the major tasks and classes. Moreover, the suitable hyperparameter range is often distinct across tabular data when applying supervised loss. The same set of hyperparameters can cause overfitting on one dataset and underfitting on another. Therefore, it is tricky to pick appropriate hyperparameters for pretraining based on vanilla supervised loss.

In this paper, we propose \clmethod for pretraining inspired by supervised CL \cite{khosla2020supervised} which was proved robust to noise and hyperparameters. As illustrated by Fig. \ref{fig:3}, we build positive pairs considering views from the same class except for only from the same sample:
\begin{equation}\label{eq:mscl}
    \ell(\bX,\by) = -\sum_{i=1}^B \sum_{j=1}^B\sum_{k=1}^K \sum_{k^\prime=1}^K \1\{y_j=y_i\}  \log \frac{\exp \psi(\bv_i^k, \bv_j^{k^\prime})}{\sum_{j^\dag=1}^B \sum_{k^\dag=1}^K \1\{y_{j^\dag}\neq y_i\} \exp \psi(\bv_i^k, \bv_{j^\dag}^{k^\dag})  }.
\end{equation}
$\by=\{y_i\}_{i}^B$ are labels; $\1\{\cdot\}$ is indicator function. \clmethod relieves multiple pretraining predictors required to adjust to different datasets. Moreover, \clmethod exposes more feature embeddings to the supervision by partitioning hence providing more discriminative and generalizable representations.

\begin{table}[t]
  \centering 
  \caption{Statistics of the use clinical trial mortality prediction datasets. All are binary classification tasks. Positive ratio means the ratio of data points belong the positive class. NCTxxx are trial identifiers which can be linked to trials on \href{https://ClinicalTrials.gov}{ClinicalTrials.gov}.}
    \begin{tabular}{llllll}
    \toprule
    Name  & Datapoints & Categorical & Binary & Numerical & Positive ratio \\
    \midrule
    NCT00041119 & 3871  & 5     & 8     & 2     & 0.07 \\
    NCT00174655 & 994   & 3     & 31    & 15    & 0.02 \\
    NCT00312208 & 1651  & 5     & 12    & 6     & 0.19 \\
    NCT00079274 & 2968  & 5     & 8     & 3     & 0.12 \\
    NCT00694382 & 1604  & 1     & 29    & 11    & 0.45 \\
    \bottomrule
    \end{tabular}%
  \label{tab:trial_data_stats}%
\end{table}%

\section{Experiments}\label{sec:experiment}
In this section, we aim at answering the following questions by extensive experiments:
\begin{itemize}[leftmargin=*, itemsep=0pt, labelsep=5pt]
    \item \textbf{Q1.} How does \method perform compared with baselines under the vanilla supervised setting?
    \item \textbf{Q2.} How well does \method address incremental columns from a stream of data (S(2) in Fig. \ref{fig:1})?
    \item \textbf{Q3.} How is the impact of \method learned from multiple tables (with different columns) drawn from the same domain on its predictive ability (S(1) in Fig. \ref{fig:1})?
    \item \textbf{Q4.} Can \method be a zero-shot learner when pretrained on tables and infer on a new table (S(4) in Fig. \ref{fig:1})?
    \item \textbf{Q5.} Is the proposed vertical partition CL better than vanilla supervised pretraining and self-supervised CL (S(3) in Fig. \ref{fig:1})?
\end{itemize}

\textbf{Datasets.} We introduce \textit{clinical trial mortality prediction datasets} where each includes a distinct group of patients and columns \footnote{\url{https://data.projectdatasphere.org/projectdatasphere/html/access}}. The data statistics are in Table \ref{tab:trial_data_stats}. Accurately predicting the patient mortality in clinical trials is crucial because it helps identify catastrophic treatment then save patients from harm and improve the clinical trial design. Considering they are from a similar domain, we can utilize them to test if \method can achieve transfer learning. Besides, we also include a set of public tabular datasets, the statistics are in Table \ref{tab:appx_data}.

\textbf{Dataset pre-processing.} For all baselines, we represent categorical features by ordinal encoding if they need to specify categorical features, otherwise one-hot encoding is used. Numerical features are scaled to $[0,1]$ by min-max normalization. Exceptionally for \method, we map the categorical feature index to its original description, e.g., mapping class "1" under "gender" to "female".

\textbf{Model and implementation protocols.} Unless specified otherwise, we keep the settings fixed across all experiments. \method uses 2 layers of gated transformers where the embedding dimensions of numbers and tokens are 128, and the hidden dimension of intermediate dense layers is 256. The attention module has 8 heads. We choose ReLU activations and do not activate dropout. We train \method using Adam optimizer \cite{kingma2014adam} with learning rate in $\{2\text{e-5},5\text{e-5},1\text{e-4}\}$ and no weight decay; batch size is in $\{16,64,128\}$. We set a maximum self-supervised pretraining epochs of 50 and supervised training epochs of 100. A patience of 10 is kept for supervised training for early stopping. Experiments were conducted with one RTX3070 GPU, i7-10700 CPU, and 16GB RAM.

\begin{table}[t]
  \centering
  \caption{Test AUROC results on clinical trial mortality datasets the under \textbf{supervised learning} setting. All the remaining tables in this paper follow these setups to avoid clutter: the metric values are averaged over 10 random seeds; the \textit{Rank} column reports the average rank across all datasets; Top results for each dataset are in bold.  } 
  \resizebox{\textwidth}{!}{%
    \begin{tabular}{llllll|l}
    \toprule
    Methods & N00041119 & N00174655 & N00312208 & N00079274 & N00694382 & Rank(Std) \\
    \midrule
    LR    & 0.6364 & 0.8543 & 0.7382 & 0.7067 & 0.7360 & 5.40(1.14) \\
    XGBoost & 0.5937 & 0.5000 & 0.6911 & 0.6784 & 0.7440 & 9.60(3.71) \\
    MLP   & 0.6340 & 0.6189 & 0.7427 & 0.6967 & 0.7063 & 8.00(2.83) \\
    SNN   & 0.6335 & 0.9130 & 0.7469 & 0.6948 & 0.7246 & 5.80(2.39) \\
    TabNet & 0.5856 & 0.5401 & 0.6910 & 0.6031 & 0.7113 & 11.40(0.89) \\
    \midrule
    DCN   & 0.6349 & 0.7577 & 0.7431 & 0.6952 & 0.7458 & 5.60(2.51) \\
    AutoInt & 0.6327 & 0.7502 & 0.7479 & 0.6958 & 0.7411 & 6.20(2.59) \\
    \midrule
    TabTrans & 0.6187 & 0.9035 & 0.7069 & 0.7178 & 0.7229 & 7.20(3.56) \\
    FT-Trans & 0.6372 & 0.9073 & 0.7586 & 0.7090 & 0.7231 & 4.20(2.28) \\
    \midrule
    VIME  & 0.6397 & 0.8533 & 0.7227 & 0.6790 & 0.7232 & 7.00(3.08) \\
    SCARF & 0.6248 & 0.9310 & 0.7267 & 0.7176 & 0.7103 & 6.60(3.91) \\
    \midrule
    \method & \textbf{0.6408} & \textbf{0.9428} & \textbf{0.7770} & \textbf{0.7281} & \textbf{0.7648} & \textbf{1.00(0.00)} \\
    \bottomrule
    \end{tabular}%
    }
  \label{tab:supervise_trial}%
\end{table}%

\begin{table}[t]
  \centering
  \caption{Test AUROC results on clinical trial datasets under \textbf{feature incremental learning}.}
 \resizebox{\textwidth}{!}{%
    \begin{tabular}{llllll|l}
    \toprule
    Methods & N00041119 & N00174655 & N00312208 & N00079274 & N00694382 & Rank(Std) \\
    \midrule
    LR    & 0.6213 & 0.8485 & 0.6801 & 0.6258 & 0.7236 & 4.6(3.21) \\
    XGBoost & 0.5735 & 0.7890 & 0.6760 & 0.6038 & 0.6463 & 8.8(2.59) \\
    MLP   & 0.6371 & 0.7754 & 0.6871 & 0.6220 & 0.6851 & 6.2(2.95) \\
    SNN   & 0.5765 & 0.7440 & 0.6854 & 0.6336 & 0.7035 & 6.4(2.30) \\
    TabNet & 0.5548 & 0.8419 & 0.5849 & 0.6052 & 0.6668 & 9.0(3.39) \\
    \midrule
    DCN   & 0.5172 & 0.5846 & 0.6640 & 0.6535 & 0.6957 & 8.2(4.16) \\
    AutoInt & 0.5232 & 0.6075 & 0.7031 & 0.6394 & 0.6974 & 7.2(3.56) \\
    \midrule
    TabTrans & 0.5599 & 0.7652 & 0.6433 & 0.6365 & 0.6841 & 8.2(1.10) \\
    FT-Trans & 0.5552 & 0.8045 & 0.7148 & 0.6471 & 0.6815 & 5.8(3.11) \\
    \midrule
    VIME  & 0.6101 & 0.8114 & 0.3705 & 0.6444 & 0.6436 & 7.4(4.22) \\
    SCARF & 0.5996 & 0.6261 & 0.7072 & 0.6535 & 0.6957 & 5.2(2.97) \\
    \midrule
    \method & \textbf{0.6797} & \textbf{0.8545} & \textbf{0.7617} & \textbf{0.6857} & \textbf{0.7795} & \textbf{1.0(0.00)} \\
    \bottomrule
    \end{tabular}%
    }
  \label{tab:incremental_trial}%
\end{table}%

\textbf{Baselines.} We include the following baselines for comparison: \textit{Logistic regression (LR)}; \textit{XGBoost} \cite{chen2016xgboost}; \textit{Multilayer perceptron (MLP)}; \textit{SeLU MLP (SNN)} \cite{klambauer2017self}; \textit{TabNet} \cite{arik2021tabnet};
 \textit{DCN} \cite{wang2017deep};
 \textit{AutoInt} \cite{song2019autoint};
 \textit{TabTransformer} \cite{huang2020tabtransformer};
 \textit{FT-Transformer} \cite{gorishniy2021revisiting};
 \textit{VIME} \cite{yoon2020vime};
 \textit{SCARF} \cite{bahri2022scarf}.
We provide the baseline architectures and implementations in Appendix \ref{appx:baseline}.

\subsection{Q1. Supervised learning}
Results of supervised learning on clinical trial mortality prediction datasets are summarized by Table \ref{tab:supervise_trial}.  Note that all methods including ours do not perform pre-training. We see that our method outperforms baselines on all. From the view of method ranks, we surprisingly identify that LR wins over half of baseline methods. Except for \method, FT-Transformer is the only model that shows significant superiority over LR, which illustrates the potential of transformers for tabular modeling.  Additional results on public datasets are available in Table \ref{tab:sup_open} where we witness that our method is comparable to the state-of-the-art baseline tabular models.  We also discover the baselines drawn from the CTR prediction literature (DCN and AutoInt) turn out the be competitive in tabular modeling. 

\subsection{Q2. Feature incremental learning} \label{sec:fil}
For previous tabular models, we should either drop new features or drop old data when confronting feature incremental learning. By contrast, \method is able to continually learn from new data with incremental features. We split the raw dataset into three subsets: set1, 2, and 3 which mimic the incremental feature scenario shown by (2) in Fig. \ref{fig:1}. Baseline methods apply to two scenarios: (a) learning from all data that only have features of set1 and (b) learning from the data of set3 only. We report the best of the two. \method applies to learning from all three subsets. Table \ref{tab:incremental_trial} shows the results where we find our method outperforms baselines by a great margin. It demonstrates that \method makes the best of incremental features to learn better. Similar observations appear in public datasets, shown by Table \ref{tab:incremental_open}.

\begin{table}[t]
  \centering
  \caption{Test AUROC results on clinical trial datasets under \textbf{transfer learning} across tables. }
   \resizebox{\textwidth}{!}{%
    \begin{tabular}{lcccccccccc|l}
    \toprule
    Methods & \multicolumn{2}{c}{N00041119} & \multicolumn{2}{c}{N00174655} & \multicolumn{2}{c}{N00312208} & \multicolumn{2}{c}{N00079274} & \multicolumn{2}{c|}{N00694382} & Rank(Std) \\
          & set1  & set2  & set1  & set2  & set1  & set2  & set1  & set2  & set1  & set2  &  \\
          \midrule
    LR    & 0.625 & 0.647 & 0.789 & 0.819 & 0.701 & 0.735 & 0.635 & 0.685 & 0.675 & 0.763 & 5.33(1.73) \\
    XGBoost & 0.638 & 0.575 & 0.574 & \textbf{0.886} & 0.690 & 0.700 & 0.596 & 0.647 & 0.592 & 0.677 & 7.56(3.75) \\
    MLP   & 0.639 & 0.621 & 0.314 & 0.857 & 0.683 & 0.744 & 0.620 & 0.675 & 0.648 & 0.765 & 6.56(3.32) \\
    SNN   & 0.627 & 0.634 & 0.215 & 0.754 & 0.687 & 0.732 & 0.631 & 0.683 & 0.651 & 0.759 & 7.44(2.07) \\
    TabNet & 0.564 & 0.558 & 0.856 & 0.592 & 0.671 & 0.657 & 0.443 & 0.605 & 0.581 & 0.677 & 10.67(2.96) \\
              \midrule
    DCN   & 0.636 & 0.625 & 0.767 & 0.790 & 0.711 & 0.698 & 0.682 & 0.664 & 0.658 & 0.737 & 6.33(2.45) \\
    AutoInt & 0.629 & 0.630 & 0.843 & 0.730 & 0.725 & 0.698 & 0.679 & 0.665 & 0.686 & 0.661 & 5.89(2.89) \\
          \midrule
    TabTrans & 0.616 & 0.647 & 0.866 & 0.822 & 0.675 & 0.677 & 0.618 & 0.702 & 0.652 & 0.718 & 6.22(3.38) \\
    FT-Trans & 0.627 & 0.641 & 0.836 & 0.858 & 0.720 & 0.741 & \textbf{0.692} & 0.692 & 0.652 & 0.740 & 4.22(2.28) \\
          \midrule
    VIME  & 0.603 & 0.625 & 0.312 & 0.726 & 0.601 & 0.642 & 0.477 & 0.668 & 0.614 & 0.715 & 10.44(1.51) \\
    SCARF & 0.635 & \textbf{0.657} & 0.651 & 0.814 & 0.653 & 0.686 & 0.682 & 0.701 & 0.671 & \textbf{0.776} & 5.56(3.40) \\
          \midrule
    \method & \textbf{0.653} & 0.653 & \textbf{0.904} & 0.846 & \textbf{0.730} & \textbf{0.756} & 0.680 & \textbf{0.711} & \textbf{0.747} & 0.774 & \textbf{1.78(1.30)} \\
    \bottomrule
    \end{tabular}%
    }
  \label{tab:transfer_trial}%
\end{table}%

\subsection{Q3. Transfer learning}\label{sec:tl}
We further test if \method is able to transfer knowledge across tables. Results are in Table \ref{tab:transfer_trial}. We split each dataset into two subsets with 50\% overlaps of their columns. Baselines are trained and tested on set1 (only label-supervision) or set2 separately. For our method we pretrain it on set1 then finetune it on set2 and report its performance on set2, and vice versa. We observe that \method can benefit from knowledge transfer across tables to reach superior performances. Similar observations are made on public datasets shown by Table \ref{tab:transfer_open}.

\begin{table}[t]
  \centering
  \caption{Test AUROC results on clinical trial datasets under \textbf{zero-shot learning} setting.}
    \begin{tabular}{llllll}
    \toprule
    \method & N00041119 & N00174655 & N00312208 & N00079274 & N00694382 \\
    \midrule
    Supervised   & 0.5854 & 0.6484 & 0.7536 & 0.7087 & 0.6479 \\
    Transfer & 0.6130 & 0.6909 & 0.7658 & 0.7163 & 0.6752 \\
    Zero-shot & 0.5990 & 0.6752 & 0.7576 & 0.7036 & 0.6740 \\
    \bottomrule
    \end{tabular}%
  \label{tab:zeroshot_trial}%
\end{table}%

\subsection{Q4. Zero-shot learning}\label{sec:zsl}
Although there are numerous papers on zero-shot learning (ZSL) in CV and NLP \cite{romera2015embarrassingly,brown2020language,radford2021learning}, we notice that ZSL was hardly mentioned in tabular domain. In this experiment, we refer to the ZSL scenario mentioned by S(4) of Fig. \ref{fig:1} where we split the raw table into three equal-size subsets. Three subsets have distinct columns. For the \textit{zero-shot} setting, the model learns from set1+set2 and is tested on set3 without further training. In this scenario, the model needs to leverage the learned knowledge from set1 and set2 to support the inference on a new table set3. Besides, we design two baselines for comparison: \textit{supervised} where the model learns from set3 and predicts on set3 and \textit{transfer} where the model learns from set1+set2 and continues to be finetuned on set3. Results are in Table \ref{tab:zeroshot_trial}. We surprisingly find the ZSL model gets better performance than the supervised one on average. It boils down to that (1) ZSL \method succeeds to retain the learned knowledge from set1+set2 for predicting on a new table (set3) and (2) ZSL can benefit from more data (set1+set2) than the supervised (set3 only). Meanwhile, the transfer model takes the advantage of set1+set2 and is adapted for set3 by finetuning, hence reaches the best performance. Similarly, we witness that \method is able to make zero-shot predictions on public datasets as in Table \ref{tab:zeroshot_open}. 

Additional sensitivity check is provided by Fig. \ref{fig:zsl_sensitivity} where we vary the overlap ratio of two subsets from the same dataset. We witness that our model makes reasonable predictions even if the training set has no column overlap with the test set.

\begin{table}[t]
  \centering
  \caption{Test AUROC on clinical trial datasets under the \textbf{across-table pretraining plus finetuning} setting. \textit{Supervised}: baseline supervised model; \textit{Transfer}: vanilla supervised transfer learning. Red shows the one worse than the \textit{Supervised} baseline.}
    \begin{tabular}{llllll}
    \toprule
    \method & N00041119 & N00174655 & N00312208 & N00079274 & N00694382 \\
    \midrule
    Supervised    & 0.6313 & 0.8348 & 0.7444 & 0.6885 & 0.7293 \\
    Transfer   & \textbf{0.6424} & \textcolor[rgb]{ 1,  0,  0}{0.8183} & 0.7458 & 0.6928 & \textcolor[rgb]{ 1,  0,  0}{0.7239} \\
    \midrule
    Self-\clmethod & 0.6412 & 0.8577 & 0.7486 & \textbf{0.7069} & 0.7348 \\
    \midrule
    \clmethod & 0.6405 & \textbf{0.8583} & \textbf{0.7517} & 0.7063 & \textbf{0.7392} \\
    \bottomrule
    \end{tabular}%
  \label{tab:csscl}%
\end{table}%

\subsection{Q5. Supervised and self-supervised pretraining}\label{sec:exp_rq5}
We take experiments to compare the proposed \clmethod with the vanilla transfer learning strategy, as in Table \ref{tab:csscl}. We observe that the vanilla strategy harms the performance on two datasets while \clmethod always brings positive effect for finetuning. Besides, we conduct experiments on varying the number of partitions and show the average AUROC on all five datasets, shown by Fig. \ref{fig:ablation_n_partition_trial}. We specify that \clmethod demonstrates an advantage over self-\clmethod when we increase the partition numbers.

We also explore if pretraining works on public datasets. Results in Table \ref{tab:pretrain_open} somewhat match our expectations that pretraining on unrelated tabular data usually yields few benefits for finetuning because these tables define totally different columns and targeted tasks. We also show the ablation on the number of partitions by Fig. \ref{fig:ablation_n_partition_open} where \clmethod consistently outperforms the \textit{Supervised} baseline. Nevertheless,  it is still worth investigating the table phenotypes to aggregate tables which are more likely to benefit from each other by transfer learning.

\section{Related Works}

\textbf{Tabular Prediction.} To enhance tabular predictions, numerous recent works try to design new algorithms \cite{chen2016xgboost,ke2017lightgbm,dorogush2018catboost,arik2021tabnet,chen2021danets,gorishniy2021revisiting,huang2020tabtransformer,somepalli2021saint,cholakov2022gatedtabtransformer,borisov2021deep,abutbul2020dnf,katzir2020net,luo2020network,guo2021tabgnn,luo2019autocross,qin2021retrieval}. However, it was argued that boosting algorithms and MLPs are still the competitive choices for tabular data modeling, especially when the sample size is small \cite{gorishniy2021revisiting,shwartz2022tabular,borisov2021deep,kadra2021well}. To alleviate label scarcity issue, SSL pretraining on unlabeled tabular data was introduced \cite{yoon2020vime,darabi2021contrastive,somepalli2021saint,ucar2021subtab,bahri2022scarf}. Nonetheless, none of them is \textit{transferable} across tables then is able to extend the success of pretraining to the tabular domain. For practical tabular predictions, the common case is that we own a lot of labeled samples collected with diverse protocols hence heavy preprocessing is needed to align them by either dropping many samples or many features. By contrast, \method accepts variable-column tables and therefore can learn from different tables at scale and transfer to the target task. Also, it can support diverse tabular prediction tasks as depicted in Fig. \ref{fig:1}, which cannot be done by off-the-shelf tabular methods.

\textbf{Transfer learning.} Transfer learning (TL) has long been a popular research field since the proposal of ImageNet \cite{deng2009imagenet}, which gives rise to splendid works on utilizing supervised pretraining on a large general database and finetune on a small downstream task \cite{simonyan2014very,yosinski2014transferable,he2016deep,huang2017densely,zoph2018learning}. TL is also fast-growing in NLP beginning at BERT \cite{devlin2018bert}, which often leverages web-scale unlabeled texts for self-supervised pretraining and then applies to specific tasks \cite{brown2020language,liu2019roberta,raffel2020exploring,yang2019xlnet,lewis2020bart}. However, few work was on TL in tabular predictions. As mentioned in \S \ref{sec:introduction}, \method paves the way for effective tabular TL by establishing a feature processing protocol that applies for most table inputs, such that it shares knowledge across tables.

\textbf{Self-supervised learning \& contrastive learning.} SSL uses unlabeled data with pretext tasks to learn useful representations and most of them are in CV and NLP \cite{devlin2018bert,bao2021beit,he2021masked,dosovitskiy2020image,gao2021simcse,vaswani2017attention,kenton2019bert,tsai2019multimodal,akbari2021vatt,lepikhin2020gshard,fedus2021switch}. Recent SSL tabular models can be classified into \textit{reconstruction} and \textit{contrastive} based methods: TabNet \cite{arik2021tabnet} and VIME \cite{yoon2020vime} try to recover the corrupted inputs with auto-encoding loss; SCARF \cite{bahri2022scarf} takes a SimCLR-like \cite{chen2020simple} contrastive loss between the sample and its corrupted version; SubTab \cite{ucar2021subtab} takes a combination of both. Nevertheless, all fail to learn transferable models across tables such that cannot benefit from pretraining with scale. Contrastive learning can also be applied to supervised learning by leveraging class labels to build positive samples \cite{khosla2020supervised}. Our work extends it to to the tabular domain, which we prove works better than vanilla supervised pretraining. The vertical partition sampling also enjoys high query speed from large databases which are often column-oriented \cite{stonebraker2018c}. Another line of research takes table pretraining table semantic parsing \cite{yin2020tabert,iida2021tabbie,lin2020bridging,deng2021turl,yang2022tableformer} or table-to-text generation \cite{wang2021retrieving,wang2022robust}. But these methods either encode the whole table instead of each row or do not demonstrate to benefit tabular prediction yet.


\section{Conclusion}
This paper proposes \method that accepts variable-column inputs. By the proposed vertical partition contrastive learning, it can benefit from supervised pretraining from multiple tabular datasets with low memory cost. We envision it to be the basis of tabular foundation models and widely used to tabular-related applications in the future. 

\section*{Acknowledgement}
This work was supported by NSF award SCH-2205289, SCH-2014438, IIS-1838042, NIH award R01 1R01NS107291-01.

{\small
\bibliography{main}

\begin{thebibliography}{10}

\bibitem{wang2017deep}
Ruoxi Wang, Bin Fu, Gang Fu, and Mingliang Wang.
\newblock Deep \& cross network for ad click predictions.
\newblock In {\em Proceedings of the ADKDD'17}, pages 1--7. 2017.

\bibitem{yoon2020vime}
Jinsung Yoon, Yao Zhang, James Jordon, and Mihaela van~der Schaar.
\newblock {VIME}: Extending the success of self-and semi-supervised learning to
  tabular domain.
\newblock {\em Advances in Neural Information Processing Systems},
  33:11033--11043, 2020.

\bibitem{zhang2020customer}
Yixuan Zhang, Jialiang Tong, Ziyi Wang, and Fengqiang Gao.
\newblock Customer transaction fraud detection using xgboost model.
\newblock In {\em International Conference on Computer Engineering and
  Application}, pages 554--558. IEEE, 2020.

\bibitem{wang2020data}
Zifeng Wang and Suzhen Li.
\newblock Data-driven risk assessment on urban pipeline network based on a
  cluster model.
\newblock {\em Reliability Engineering \& System Safety}, 196:106781, 2020.

\bibitem{huang2020tabtransformer}
Xin Huang, Ashish Khetan, Milan Cvitkovic, and Zohar Karnin.
\newblock {TabTransformer}: Tabular data modeling using contextual embeddings.
\newblock {\em arXiv preprint arXiv:2012.06678}, 2020.

\bibitem{padhi2021tabular}
Inkit Padhi, Yair Schiff, Igor Melnyk, Mattia Rigotti, Youssef Mroueh, Pierre
  Dognin, Jerret Ross, Ravi Nair, and Erik Altman.
\newblock Tabular transformers for modeling multivariate time series.
\newblock In {\em IEEE International Conference on Acoustics, Speech and Signal
  Processing}, pages 3565--3569. IEEE, 2021.

\bibitem{cholakov2022gatedtabtransformer}
Radostin Cholakov and Todor Kolev.
\newblock The {GatedTabTransformer}. an enhanced deep learning architecture for
  tabular modeling.
\newblock {\em arXiv preprint arXiv:2201.00199}, 2022.

\bibitem{wang2021survtrace}
Zifeng Wang and Jimeng Sun.
\newblock {SurvTRACE}: Transformers for survival analysis with competing
  events.
\newblock {\em arXiv preprint arXiv:2110.00855}, 2021.

\bibitem{ucar2021subtab}
Talip Ucar, Ehsan Hajiramezanali, and Lindsay Edwards.
\newblock {SubTab}: Subsetting features of tabular data for self-supervised
  representation learning.
\newblock {\em Advances in Neural Information Processing Systems}, 34, 2021.

\bibitem{somepalli2021saint}
Gowthami Somepalli, Micah Goldblum, Avi Schwarzschild, C~Bayan Bruss, and Tom
  Goldstein.
\newblock {SAINT}: Improved neural networks for tabular data via row attention
  and contrastive pre-training.
\newblock {\em arXiv preprint arXiv:2106.01342}, 2021.

\bibitem{bahri2022scarf}
Dara Bahri, Heinrich Jiang, Yi~Tay, and Donald Metzler.
\newblock {SCARF}: Self-supervised contrastive learning using random feature
  corruption.
\newblock In {\em International Conference on Learning Representations}, 2022.

\bibitem{nargesian2018table}
Fatemeh Nargesian, Erkang Zhu, Ken~Q Pu, and Ren{\'e}e~J Miller.
\newblock Table union search on open data.
\newblock {\em Proceedings of the VLDB Endowment}, 11(7):813--825, 2018.

\bibitem{zhu2019josie}
Erkang Zhu, Dong Deng, Fatemeh Nargesian, and Ren{\'e}e~J Miller.
\newblock Josie: Overlap set similarity search for finding joinable tables in
  data lakes.
\newblock In {\em International Conference on Management of Data}, pages
  847--864, 2019.

\bibitem{dong2021efficient}
Yuyang Dong, Kunihiro Takeoka, Chuan Xiao, and Masafumi Oyamada.
\newblock Efficient joinable table discovery in data lakes: A high-dimensional
  similarity-based approach.
\newblock In {\em IEEE International Conference on Data Engineering}, pages
  456--467. IEEE, 2021.

\bibitem{he2021masked}
Kaiming He, Xinlei Chen, Saining Xie, Yanghao Li, Piotr Doll{\'a}r, and Ross
  Girshick.
\newblock Masked autoencoders are scalable vision learners.
\newblock {\em arXiv preprint arXiv:2111.06377}, 2021.

\bibitem{dosovitskiy2020image}
Alexey Dosovitskiy, Lucas Beyer, Alexander Kolesnikov, Dirk Weissenborn,
  Xiaohua Zhai, Thomas Unterthiner, Mostafa Dehghani, Matthias Minderer, Georg
  Heigold, Sylvain Gelly, et~al.
\newblock An image is worth 16x16 words: Transformers for image recognition at
  scale.
\newblock In {\em International Conference on Learning Representations}, 2020.

\bibitem{bao2021beit}
Hangbo Bao, Li~Dong, and Furu Wei.
\newblock {BEiT}: Bert pre-training of image transformers.
\newblock {\em arXiv preprint arXiv:2106.08254}, 2021.

\bibitem{mikolov2013efficient}
Tomas Mikolov, Kai Chen, Greg Corrado, and Jeffrey Dean.
\newblock Efficient estimation of word representations in vector space.
\newblock {\em arXiv preprint arXiv:1301.3781}, 2013.

\bibitem{schuster2012japanese}
Mike Schuster and Kaisuke Nakajima.
\newblock Japanese and korean voice search.
\newblock In {\em IEEE international conference on acoustics, speech and signal
  processing (ICASSP)}, pages 5149--5152. IEEE, 2012.

\bibitem{devlin2018bert}
Jacob Devlin, Ming-Wei Chang, Kenton Lee, and Kristina Toutanova.
\newblock Bert: Pre-training of deep bidirectional transformers for language
  understanding.
\newblock {\em arXiv preprint arXiv:1810.04805}, 2018.

\bibitem{lin2020birds}
Bill~Yuchen Lin, Seyeon Lee, Rahul Khanna, and Xiang Ren.
\newblock Birds have four legs?! {NumerSense}: Probing numerical commonsense
  knowledge of pre-trained language models.
\newblock In {\em Proceedings of the EMNLP}, pages 6862--6868, 2020.

\bibitem{ba2016layer}
Jimmy~Lei Ba, Jamie~Ryan Kiros, and Geoffrey~E Hinton.
\newblock Layer normalization.
\newblock {\em arXiv preprint arXiv:1607.06450}, 2016.

\bibitem{vaswani2017attention}
Ashish Vaswani, Noam Shazeer, Niki Parmar, Jakob Uszkoreit, Llion Jones,
  Aidan~N Gomez, {\L}ukasz Kaiser, and Illia Polosukhin.
\newblock Attention is all you need.
\newblock {\em Advances in Neural Information Processing Systems}, 30, 2017.

\bibitem{darabi2021contrastive}
Sajad Darabi, Shayan Fazeli, Ali Pazoki, Sriram Sankararaman, and Majid
  Sarrafzadeh.
\newblock Contrastive mixup: Self-and semi-supervised learning for tabular
  domain.
\newblock {\em arXiv preprint arXiv:2108.12296}, 2021.

\bibitem{stonebraker2018c}
Mike Stonebraker, Daniel~J Abadi, Adam Batkin, Xuedong Chen, Mitch Cherniack,
  Miguel Ferreira, Edmond Lau, Amerson Lin, Sam Madden, Elizabeth O'Neil,
  et~al.
\newblock C-store: a column-oriented dbms.
\newblock In {\em Making Databases Work: the Pragmatic Wisdom of Michael
  Stonebraker}, pages 491--518. 2018.

\bibitem{khosla2020supervised}
Prannay Khosla, Piotr Teterwak, Chen Wang, Aaron Sarna, Yonglong Tian, Phillip
  Isola, Aaron Maschinot, Ce~Liu, and Dilip Krishnan.
\newblock Supervised contrastive learning.
\newblock {\em Advances in Neural Information Processing Systems},
  33:18661--18673, 2020.

\bibitem{kingma2014adam}
Diederik~P Kingma and Jimmy Ba.
\newblock Adam: A method for stochastic optimization.
\newblock {\em arXiv preprint arXiv:1412.6980}, 2014.

\bibitem{chen2016xgboost}
Tianqi Chen and Carlos Guestrin.
\newblock {Xgboost}: A scalable tree boosting system.
\newblock In {\em ACM SIGKDD International Conference on Knowledge Discovery
  and Data Mining}, pages 785--794, 2016.

\bibitem{klambauer2017self}
G{\"u}nter Klambauer, Thomas Unterthiner, Andreas Mayr, and Sepp Hochreiter.
\newblock Self-normalizing neural networks.
\newblock {\em Advances in Neural Information Processing Systems}, 30, 2017.

\bibitem{arik2021tabnet}
Sercan~O Ar{\i}k and Tomas Pfister.
\newblock Tabnet: Attentive interpretable tabular learning.
\newblock In {\em AAAI}, volume~35, pages 6679--6687, 2021.

\bibitem{song2019autoint}
Weiping Song, Chence Shi, Zhiping Xiao, Zhijian Duan, Yewen Xu, Ming Zhang, and
  Jian Tang.
\newblock {AutoInt}: Automatic feature interaction learning via self-attentive
  neural networks.
\newblock In {\em Proceedings of the 28th ACM International Conference on
  Information and Knowledge Management}, pages 1161--1170, 2019.

\bibitem{gorishniy2021revisiting}
Yury Gorishniy, Ivan Rubachev, Valentin Khrulkov, and Artem Babenko.
\newblock Revisiting deep learning models for tabular data.
\newblock {\em Advances in Neural Information Processing Systems}, 34, 2021.

\bibitem{romera2015embarrassingly}
Bernardino Romera-Paredes and Philip Torr.
\newblock An embarrassingly simple approach to zero-shot learning.
\newblock In {\em International conference on machine learning}, pages
  2152--2161. PMLR, 2015.

\bibitem{brown2020language}
Tom Brown, Benjamin Mann, Nick Ryder, Melanie Subbiah, Jared~D Kaplan, Prafulla
  Dhariwal, Arvind Neelakantan, Pranav Shyam, Girish Sastry, Amanda Askell,
  et~al.
\newblock Language models are few-shot learners.
\newblock {\em Advances in Neural Information Processing Systems},
  33:1877--1901, 2020.

\bibitem{radford2021learning}
Alec Radford, Jong~Wook Kim, Chris Hallacy, Aditya Ramesh, Gabriel Goh,
  Sandhini Agarwal, Girish Sastry, Amanda Askell, Pamela Mishkin, Jack Clark,
  et~al.
\newblock Learning transferable visual models from natural language
  supervision.
\newblock In {\em International Conference on Machine Learning}, pages
  8748--8763. PMLR, 2021.

\bibitem{ke2017lightgbm}
Guolin Ke, Qi~Meng, Thomas Finley, Taifeng Wang, Wei Chen, Weidong Ma, Qiwei
  Ye, and Tie-Yan Liu.
\newblock {LightGBM}: A highly efficient gradient boosting decision tree.
\newblock {\em Advances in Neural Information Processing Systems}, 30, 2017.

\bibitem{dorogush2018catboost}
Anna~Veronika Dorogush, Vasily Ershov, and Andrey Gulin.
\newblock {CatBoost}: gradient boosting with categorical features support.
\newblock {\em arXiv preprint arXiv:1810.11363}, 2018.

\bibitem{chen2021danets}
Jintai Chen, Kuanlun Liao, Yao Wan, Danny~Z Chen, and Jian Wu.
\newblock Danets: Deep abstract networks for tabular data classification and
  regression.
\newblock {\em arXiv preprint arXiv:2112.02962}, 2021.

\bibitem{borisov2021deep}
Vadim Borisov, Tobias Leemann, Kathrin Se{\ss}ler, Johannes Haug, Martin
  Pawelczyk, and Gjergji Kasneci.
\newblock Deep neural networks and tabular data: A survey.
\newblock {\em arXiv preprint arXiv:2110.01889}, 2021.

\bibitem{abutbul2020dnf}
Ami Abutbul, Gal Elidan, Liran Katzir, and Ran El-Yaniv.
\newblock Dnf-net: A neural architecture for tabular data.
\newblock {\em arXiv preprint arXiv:2006.06465}, 2020.

\bibitem{katzir2020net}
Liran Katzir, Gal Elidan, and Ran El-Yaniv.
\newblock Net-dnf: Effective deep modeling of tabular data.
\newblock In {\em International Conference on Learning Representations}, 2020.

\bibitem{luo2020network}
Yuanfei Luo, Hao Zhou, Wei-Wei Tu, Yuqiang Chen, Wenyuan Dai, and Qiang Yang.
\newblock Network on network for tabular data classification in real-world
  applications.
\newblock In {\em Proceedings of the 43rd International ACM SIGIR Conference on
  Research and Development in Information Retrieval}, pages 2317--2326, 2020.

\bibitem{guo2021tabgnn}
Xiawei Guo, Yuhan Quan, Huan Zhao, Quanming Yao, Yong Li, and Weiwei Tu.
\newblock {TabGNN}: Multiplex graph neural network for tabular data prediction.
\newblock {\em arXiv preprint arXiv:2108.09127}, 2021.

\bibitem{luo2019autocross}
Yuanfei Luo, Mengshuo Wang, Hao Zhou, Quanming Yao, Wei-Wei Tu, Yuqiang Chen,
  Wenyuan Dai, and Qiang Yang.
\newblock Autocross: Automatic feature crossing for tabular data in real-world
  applications.
\newblock In {\em ACM SIGKDD International Conference on Knowledge Discovery \&
  Data Mining}, pages 1936--1945, 2019.

\bibitem{qin2021retrieval}
Jiarui Qin, Weinan Zhang, Rong Su, Zhirong Liu, Weiwen Liu, Ruiming Tang,
  Xiuqiang He, and Yong Yu.
\newblock Retrieval \& interaction machine for tabular data prediction.
\newblock In {\em ACM SIGKDD Conference on Knowledge Discovery \& Data Mining},
  pages 1379--1389, 2021.

\bibitem{shwartz2022tabular}
Ravid Shwartz-Ziv and Amitai Armon.
\newblock Tabular data: Deep learning is not all you need.
\newblock {\em Information Fusion}, 81:84--90, 2022.

\bibitem{kadra2021well}
Arlind Kadra, Marius Lindauer, Frank Hutter, and Josif Grabocka.
\newblock Well-tuned simple nets excel on tabular datasets.
\newblock {\em Advances in Neural Information Processing Systems}, 34, 2021.

\bibitem{deng2009imagenet}
Jia Deng, Wei Dong, Richard Socher, Li-Jia Li, Kai Li, and Li~Fei-Fei.
\newblock Imagenet: A large-scale hierarchical image database.
\newblock In {\em IEEE Conference on Computer Vision and Pattern Recognition},
  pages 248--255. IEEE, 2009.

\bibitem{simonyan2014very}
Karen Simonyan and Andrew Zisserman.
\newblock Very deep convolutional networks for large-scale image recognition.
\newblock {\em arXiv preprint arXiv:1409.1556}, 2014.

\bibitem{yosinski2014transferable}
Jason Yosinski, Jeff Clune, Yoshua Bengio, and Hod Lipson.
\newblock How transferable are features in deep neural networks?
\newblock {\em Advances in Neural Information Processing Systems}, 27, 2014.

\bibitem{he2016deep}
Kaiming He, Xiangyu Zhang, Shaoqing Ren, and Jian Sun.
\newblock Deep residual learning for image recognition.
\newblock In {\em IEEE Conference on Computer Vision and Pattern Recognition},
  pages 770--778, 2016.

\bibitem{huang2017densely}
Gao Huang, Zhuang Liu, Laurens Van Der~Maaten, and Kilian~Q Weinberger.
\newblock Densely connected convolutional networks.
\newblock In {\em IEEE Conference on Computer Vision and Pattern Recognition},
  pages 4700--4708, 2017.

\bibitem{zoph2018learning}
Barret Zoph, Vijay Vasudevan, Jonathon Shlens, and Quoc~V Le.
\newblock Learning transferable architectures for scalable image recognition.
\newblock In {\em IEEE Conference on Computer Vision and Pattern Recognition},
  pages 8697--8710, 2018.

\bibitem{liu2019roberta}
Yinhan Liu, Myle Ott, Naman Goyal, Jingfei Du, Mandar Joshi, Danqi Chen, Omer
  Levy, Mike Lewis, Luke Zettlemoyer, and Veselin Stoyanov.
\newblock Roberta: A robustly optimized bert pretraining approach.
\newblock {\em arXiv preprint arXiv:1907.11692}, 2019.

\bibitem{raffel2020exploring}
Colin Raffel, Noam Shazeer, Adam Roberts, Katherine Lee, Sharan Narang, Michael
  Matena, Yanqi Zhou, Wei Li, and Peter~J Liu.
\newblock Exploring the limits of transfer learning with a unified text-to-text
  transformer.
\newblock {\em Journal of Machine Learning Research}, 21:1--67, 2020.

\bibitem{yang2019xlnet}
Zhilin Yang, Zihang Dai, Yiming Yang, Jaime Carbonell, Russ~R Salakhutdinov,
  and Quoc~V Le.
\newblock {XLNet}: Generalized autoregressive pretraining for language
  understanding.
\newblock {\em Advances in Neural Information Processing Systems}, 32, 2019.

\bibitem{lewis2020bart}
Mike Lewis, Yinhan Liu, Naman Goyal, Marjan Ghazvininejad, Abdelrahman Mohamed,
  Omer Levy, Veselin Stoyanov, and Luke Zettlemoyer.
\newblock {BART}: Denoising sequence-to-sequence pre-training for natural
  language generation, translation, and comprehension.
\newblock In {\em Annual Meeting of the Association for Computational
  Linguistics}, pages 7871--7880, 2020.

\bibitem{gao2021simcse}
Tianyu Gao, Xingcheng Yao, and Danqi Chen.
\newblock Simcse: Simple contrastive learning of sentence embeddings.
\newblock In {\em Conference on Empirical Methods in Natural Language
  Processing}, pages 6894--6910, 2021.

\bibitem{kenton2019bert}
Jacob Devlin Ming-Wei~Chang Kenton and Lee~Kristina Toutanova.
\newblock {BERT}: Pre-training of deep bidirectional transformers for language
  understanding.
\newblock In {\em Proceedings of NAACL-HLT}, pages 4171--4186, 2019.

\bibitem{tsai2019multimodal}
Yao-Hung~Hubert Tsai, Shaojie Bai, Paul~Pu Liang, J~Zico Kolter, Louis-Philippe
  Morency, and Ruslan Salakhutdinov.
\newblock Multimodal transformer for unaligned multimodal language sequences.
\newblock In {\em Proceedings of the Annual Meeting of the Association for
  Computational Linguistics}, 2019.

\bibitem{akbari2021vatt}
Hassan Akbari, Liangzhe Yuan, Rui Qian, Wei-Hong Chuang, Shih-Fu Chang, Yin
  Cui, and Boqing Gong.
\newblock {VATT}: Transformers for multimodal self-supervised learning from raw
  video, audio and text.
\newblock {\em Advances in Neural Information Processing Systems}, 34, 2021.

\bibitem{lepikhin2020gshard}
Dmitry Lepikhin, HyoukJoong Lee, Yuanzhong Xu, Dehao Chen, Orhan Firat, Yanping
  Huang, Maxim Krikun, Noam Shazeer, and Zhifeng Chen.
\newblock {GShard}: Scaling giant models with conditional computation and
  automatic sharding.
\newblock In {\em International Conference on Learning Representations}, 2020.

\bibitem{fedus2021switch}
William Fedus, Barret Zoph, and Noam Shazeer.
\newblock Switch transformers: Scaling to trillion parameter models with simple
  and efficient sparsity.
\newblock {\em arXiv preprint arXiv:2101.03961}, 2021.

\bibitem{chen2020simple}
Ting Chen, Simon Kornblith, Mohammad Norouzi, and Geoffrey Hinton.
\newblock A simple framework for contrastive learning of visual
  representations.
\newblock In {\em International conference on machine learning}, pages
  1597--1607. PMLR, 2020.

\bibitem{yin2020tabert}
Pengcheng Yin, Graham Neubig, Wen-tau Yih, and Sebastian Riedel.
\newblock {TaBERT}: Pretraining for joint understanding of textual and tabular
  data.
\newblock In {\em Annual Meeting of the Association for Computational
  Linguistics}, pages 8413--8426, 2020.

\bibitem{iida2021tabbie}
Hiroshi Iida, Dung Thai, Varun Manjunatha, and Mohit Iyyer.
\newblock {TABBIE}: Pretrained representations of tabular data.
\newblock In {\em Conference of the North American Chapter of the Association
  for Computational Linguistics}, 2021.

\bibitem{lin2020bridging}
Xi~Victoria Lin, Richard Socher, and Caiming Xiong.
\newblock Bridging textual and tabular data for cross-domain text-to-sql
  semantic parsing.
\newblock {\em arXiv preprint arXiv:2012.12627}, 2020.

\bibitem{deng2021turl}
Xiang Deng, Huan Sun, Alyssa Lees, You Wu, and Cong Yu.
\newblock {TURL}: Table understanding through representation learning.
\newblock {\em Proceedings of the VLDB Endowment}, 2021.

\bibitem{yang2022tableformer}
Jingfeng Yang, Aditya Gupta, Shyam Upadhyay, Luheng He, Rahul Goel, and Shachi
  Paul.
\newblock {TableFormer}: Robust transformer modeling for table-text encoding.
\newblock In {\em Proceedings of the 60th Annual Meeting of the Association for
  Computational Linguistics}, pages 528--537, 2022.

\bibitem{wang2021retrieving}
Fei Wang, Kexuan Sun, Muhao Chen, Jay Pujara, and Pedro~A Szekely.
\newblock Retrieving complex tables with multi-granular graph representation
  learning.
\newblock In {\em SIGIR}, 2021.

\bibitem{wang2022robust}
Fei Wang, Zhewei Xu, Pedro Szekely, and Muhao Chen.
\newblock Robust (controlled) table-to-text generation with structure-aware
  equivariance learning.
\newblock {\em arXiv preprint arXiv:2205.03972}, 2022.

\bibitem{cutrona2019semantic}
Vincenzo Cutrona.
\newblock Semantic enrichment for large-scale data analytics.
\newblock 2019.

\bibitem{paszke2019pytorch}
Adam Paszke, Sam Gross, Francisco Massa, Adam Lerer, James Bradbury, Gregory
  Chanan, Trevor Killeen, Zeming Lin, Natalia Gimelshein, Luca Antiga, et~al.
\newblock Pytorch: An imperative style, high-performance deep learning library.
\newblock {\em Advances in Neural Information Processing Systems}, 32, 2019.

\bibitem{van2018representation}
Aaron Van~den Oord, Yazhe Li, Oriol Vinyals, et~al.
\newblock Representation learning with contrastive predictive coding.
\newblock {\em arXiv preprint arXiv:1807.03748}, 2(3):4, 2018.

\end{thebibliography}
\bibliographystyle{unsrt}
}

\section*{Checklist}


\begin{enumerate}

\item For all authors...
\begin{enumerate}
  \item Do the main claims made in the abstract and introduction accurately reflect the paper's contributions and scope?
    \answerYes{}
  \item Did you describe the limitations of your work?
    \answerYes{See \S ~\ref{sec:exp_rq5} and Appendix \S \ref{appx:broad_impact}.}
  \item Did you discuss any potential negative societal impacts of your work?
    \answerYes{See Appendix \S \ref{appx:broad_impact}.}
  \item Have you read the ethics review guidelines and ensured that your paper conforms to them?
    \answerYes{}
\end{enumerate}

\item If you are including theoretical results...
\begin{enumerate}
  \item Did you state the full set of assumptions of all theoretical results?
    \answerNA{This paper does not include theoretical results.}
        \item Did you include complete proofs of all theoretical results?
    \answerNA{This paper does not include theoretical results.}
\end{enumerate}

\item If you ran experiments...
\begin{enumerate}
  \item Did you include the code, data, and instructions needed to reproduce the main experimental results (either in the supplemental material or as a URL)?
    \answerYes{See supplementary materials.}
  \item Did you specify all the training details (e.g., data splits, hyperparameters, how they were chosen)?
    \answerYes{For our methods please see \textbf{Model and Implementation Protocols} of \S \ref{sec:experiment}; for the compared baselines please see Appendix \S \ref{appx:baseline};}
        \item Did you report error bars (e.g., with respect to the random seed after running experiments multiple times)?
    \answerYes{}
        \item Did you include the total amount of compute and the type of resources used (e.g., type of GPUs, internal cluster, or cloud provider)?
    \answerYes{Please see \textbf{Model and Implementation Protocols} of \S \ref{sec:experiment}.}
\end{enumerate}

\item If you are using existing assets (e.g., code, data, models) or curating/releasing new assets...
\begin{enumerate}
  \item If your work uses existing assets, did you cite the creators?
    \answerYes{See Table \ref{tab:data_url}.}
  \item Did you mention the license of the assets?
    \answerYes{Licenses are available referring to the provided links.}
  \item Did you include any new assets either in the supplemental material or as a URL?
    \answerYes{See Appendix \S \ref{appx:trial_data}.}
  \item Did you discuss whether and how consent was obtained from people whose data you're using/curating?
    \answerYes{See Appendix \S \ref{appx:trial_data}.}
  \item Did you discuss whether the data you are using/curating contains personally identifiable information or offensive content?
    \answerYes{See Appendix \S \ref{appx:trial_data}.}
\end{enumerate}

\item If you used crowdsourcing or conducted research with human subjects...
\begin{enumerate}
  \item Did you include the full text of instructions given to participants and screenshots, if applicable?
    \answerNA{This paper does not use crowdsourcing.}
  \item Did you describe any potential participant risks, with links to Institutional Review Board (IRB) approvals, if applicable?
    \answerNA{This paper does not use crowdsourcing.}
  \item Did you include the estimated hourly wage paid to participants and the total amount spent on participant compensation?
    \answerNA{This paper does not use crowdsourcing.}
\end{enumerate}

\end{enumerate}


\newpage

\appendix

\setcounter{theorem}{0}
\setcounter{lemma}{0}
\setcounter{equation}{0}
\setcounter{proposition}{0}
\numberwithin{equation}{section}
\numberwithin{lemma}{section}
\numberwithin{definition}{section}

\section{Broader impact of this work} \label{appx:broad_impact}
Tabular data is the most common data format used in practical data analysis including healthcare, finance, business, advertising, manufacturing, etc. Regardless of its importance, much more effort was made to play deep learning with other domains like vision, language, and audio. As a result, non-deep algorithms especially tree-based still dominate tabular data analysis. This paper opens the door to leverage the power of transfer learning in the tabular domain because \method is capable of dealing with variable-column input tables. This property also alleviates the workload required for data preprocessing because \method permits missing features and make good predictions based on the remaining features. This advance is expected to bring tremendous savings of time and money from the data engineering which often takes up 80\% of efforts for data science projects \cite{cutrona2019semantic}. It also sheds light on developing foundation models for the tabular domain because of the success of pretraining on scale.

The potential negative effect would be that it requires more effort on solving data privacy issues because \method works better when features are represented by descriptive texts compared with discretized indices. However, this can be alleviated by mapping features to machine-readable tokens in advance by a private codebook. Besides, \method needs more resources than simple models like MLP and Trees. The main cause is the use of full attention in multihead attention modules and the tokenization in the featurizing process. The former problem can be alleviated by replacing transformers with MLP-based blocks like gated attention units. The latter can be circumvented by pre-tokenization before the model training.

\section{Baseline architecture and implementation}\label{appx:baseline}
In this section, we introduce the implementation details of baselines in experiments. 
\begin{itemize}[leftmargin=*, itemsep=0pt, labelsep=5pt]
    \item LR: Use the default setting of the package scikit-learn\footnote{\href{https://scikit-learn.org/stable/modules/generated/sklearn.linear_model.LogisticRegression.html}{sklearn.linear\_model.LogisticRegression}} except the max iterations is set 1000.
    \item XGBoost: We set the maximum number of estimators in $\{50, 100, 500\}$ and the max depth in $\{4,6,8\}$. Implemented based on the XGBoost package\footnote{\href{https://xgboost.readthedocs.io/en/stable/python/python_api.html}{xgboost.sklearn}}.
    \item MLP \& SNN: We keep the same architecture for both except for their activations: three dense layers with hidden dimensions 256, 256, 1; dropout with rate of 0.1 is used. They are trained with batch size $\in \{16, 32, 64,128\}$, learning rate $\in$ \{5e-5, 1e-4, 5e-4, 1e-3\}, and early stopping patience of 5 with 100 maximum epochs.
    \item TabNet: Use the official implementation with the default recommended parameters\footnote{\url{https://github.com/dreamquark-ai/tabnet}}. Trained with batch size $\in \{16, 32, 64, 128\}$, learning rate $\in$ \{1e-4, 1e-3, 2e-2\}, $n_a,n_d \in \{8, 16, 64, 128\}$, $\gamma \in \{1.3, 1.5, 1.8\}$, categorical embedding dimension $\in \{1, 8, 16\}$ and early stopping patience of 5 with 100 maximum epochs.
    \item DCN: Use the implementation by Deep-CTR\footnote{\url{https://github.com/shenweichen/DeepCTR-Torch}}. The number of cross is 2; dropout rate for feed-forward component is 0.1; MLP part has two dense layers of dimension 256, 128; Trained with batch size $\in \{16,32,64,128\}$, learning rate $\in$ \{5e-5, 1e-4, 1e-3\}, and early stopping patience of 10 in 100 maximum epochs.
    \item AutoInt: Use the implementation by Deep-CTR. Attention layer number is set 2; Attention head number is set 2; MLP part has two dense layers of dimension 256, 128; dropout deactivated; Trained with batch size $\in \{16,32,64,128\}$, learning rate $\in$ \{5e-5, 1e-4, 1e-3\}, and early stopping patience of 10 in 100 maximum epochs.
    \item TabTransformer: Use the official implementation\footnote{\url{https://github.com/lucidrains/tab-transformer-pytorch}}. Feed-forward component has 128 dimension; 2 transformer layers are used; The number of heads of attention is $\in \{2, 4, 8\}$; Dropout rate is 0.1; ReLU activation is used; Trained with batch size $\in \{16,32,64,128\}$, learning rate $\in$ \{5e-5,1e-4,1e-3\}, and early stopping patience of 10 in 100 maximum epochs.
    \item FT-Transformer: Use the official implementation\footnote{\url{https://github.com/Yura52/rtdl}}. Feed-forward component has 128 dimension; 2 transformer layers are used; The number of heads of attention is $\in \{2,4,8\}$; Dropout rate is 0.1; ReLU activation is used; Trained with batch size $\in \{16,32,64\}$, learning rate $\in$ \{5e-5,1e-4,1e-3\}, and early stopping patience of 10 in 100 maximum epochs.
    \item VIME: We reproduce it by PyTorch \cite{paszke2019pytorch} based on the original official implementation\footnote{\url{https://github.com/jsyoon0823/VIME}} where its encoders, mask estimators, decoders are all one dense layer with the hidden dimension same as the input features. During the pretraining phase, we train the model on all training data taking mask rate 0.3, batch size 128, learning rate 1e-4, and 10 epochs; during the fine-tuning phase, we add a classifier after the encoder with three dense layers of 100 dimension and ReLU activations. Trained with batch size $\in \{16,32,64,128\}$, learning rate \{5e-5,1e-4,1e-3\}, and early stopping patience of 10 in 100 maximum epochs.
    \item SCARF: Since no official code is found, we reproduce it by PyTorch based on the descriptions in the original paper. A 4-layer encoder and a 2-layer decoder with ReLU activations are used. Hidden dimensions of their intermediate layers are all 256. During the pretraining phase, we train the model on all trainign data taking mask rate 0.5, InfoNCE loss \cite{van2018representation} with learning rate 1e-3, batch size 128, and 10 epochs; during the fine-tuning phase, we add a classifier after the encoder with two dense layers with 256 hidden dimensions. Trained with batch size $\in \{16,32,64,128\}$, learning rate $\in$ \{5e-5,1e-4,1e-3\}, and early stopping patience of 10 in 100 maximum epochs.
\end{itemize}

\begin{table}[t]
  \centering
  \caption{Statistics of the used public datasets. All are binary classification tasks. Positive ratio means the ratio of data points belong the positive class. Source links are available at Table \ref{tab:data_url}. \label{tab:appx_data}}
    \begin{tabular}{llllll}
    \toprule
    Name  & N Datapoints & Categorical & Binary & Numerical & Positive ratio \\
    \midrule
    credit-g (CG)  & 1000  & 11    & 2     & 7     & 0.7 \\
    credit-approval (CA) & 690   & 6     & 3     & 6     & 0.56 \\
    dresses-sales (DS) & 500   & 11    & 0     & 1     & 0.42 \\
    adult (AD) & 48842 & 12    & 0     & 2     & 0.24 \\
    cylinder-bands (CB) & 540   & 13    & 4     & 18    & 0.58 \\
    blastchar (BL) & 7043  & 11    & 5     & 3     & 0.27 \\
    insurance-co (IO) & 5822  & 2     & 0     & 83    & 0.06 \\
    1995-income (IC) & 32561 & 8     & 0     & 6     & 0.24 \\
    \bottomrule
    \end{tabular}%
\end{table}%

\begin{table}[t]
  \centering
  \caption{Test AUROC results on public datasets the under \textbf{supervised learning} setting.  The dataset name is abbreviated referring to Table \ref{tab:appx_data}.}
  \resizebox{\textwidth}{!}{%
    \begin{tabular}{lllllllll|l}
    \toprule
    Methods & CG    & CA    & DS    & AD    & CB    & BL    & IO    & IC    & Rank(Std) \\
    \midrule
    LR    & 0.720 & 0.836 & 0.557 & 0.851 & 0.748 & 0.801 & 0.769 & 0.860 & 9.88(1.90) \\
    XGBoost & 0.726 & \textbf{0.895} & 0.587 & 0.912 & \textbf{0.892} & 0.821 & 0.758 & \textbf{0.925} & 5.12(3.86) \\
    MLP   & 0.643 & 0.832 & 0.568 & 0.904 & 0.613 & 0.832 & 0.779 & 0.893 & 9.25(2.07) \\
    SNN   & 0.641 & 0.880 & 0.540 & 0.902 & 0.621 & 0.834 & 0.794 & 0.892 & 8.00(3.32) \\
    TabNet & 0.585 & 0.800 & 0.478 & 0.904 & 0.680 & 0.819 & 0.742 & 0.896 & 10.75(1.49) \\
    \midrule
    DCN   & 0.739 & 0.870 & \textbf{0.674} & 0.913 & 0.848 & 0.840 & 0.768 & 0.915 & 4.12(1.69) \\
    AutoInt & 0.744 & 0.866 & 0.672 & 0.913 & 0.808 & 0.844 & 0.762 & 0.916 & 4.62(2.52) \\
    \midrule
    TabTrans & 0.718 & 0.860 & 0.648 & \textbf{0.914} & 0.855 & 0.820 & 0.794 & 0.882 & 6.50(3.12) \\
    FT-Trans & 0.739 & 0.859 & 0.657 & 0.913 & 0.862 & 0.841 & 0.793 & 0.915 & 3.94(1.35) \\
    \midrule
    VIME  & 0.735 & 0.852 & 0.485 & 0.912 & 0.769 & 0.837 & 0.786 & 0.908 & 6.06(2.83) \\
    SCARF & 0.733 & 0.861 & 0.663 & 0.911 & 0.719 & 0.833 & 0.758 & 0.905 & 6.75(2.12) \\
    \midrule
    \method & \textbf{0.768} & 0.881 & 0.643 & 0.907 & 0.851 & \textbf{0.845} & \textbf{0.822} & 0.919 & \textbf{3.00(1.93)} \\
    \bottomrule
    \end{tabular}%
    }
  \label{tab:sup_open}%
\end{table}%

\begin{table}[t]
  \centering
  \caption{Test AUROC results on public datasets the under \textbf{feature incremental learning} setting. The dataset name is abbreviated referring to Table \ref{tab:appx_data}.}
    \begin{tabular}{lrrrrrrrr|l}
    \toprule
    Methods & \multicolumn{1}{l}{CG} & \multicolumn{1}{l}{CA} & \multicolumn{1}{l}{DS} & \multicolumn{1}{l}{AD} & \multicolumn{1}{l}{CB} & \multicolumn{1}{l}{BL} & \multicolumn{1}{l}{IO} & \multicolumn{1}{l}{IC} & Rank(Std) \\
    \midrule
    LR    & 0.670 & 0.773 & 0.475 & 0.832 & 0.727 & 0.806 & 0.655 & 0.825 & 8.88(2.80) \\
    XGBoost & 0.608 & 0.817 & 0.527 & 0.891 & 0.778 & 0.816 & 0.692 & \textbf{0.898} & 5.50(3.13) \\
    MLP   & 0.586 & 0.676 & 0.516 & 0.890 & 0.631 & 0.825 & 0.626 & 0.885 & 9.50(2.07) \\
    SNN   & 0.583 & 0.738 & 0.442 & 0.888 & 0.644 & 0.818 & 0.643 & 0.881 & 9.69(1.28) \\
    TabNet & 0.573 & 0.689 & 0.419 & 0.886 & 0.571 & 0.837 & 0.680 & 0.882 & 9.19(3.34) \\
    \midrule
    DCN   & 0.674 & 0.835 & 0.578 & 0.893 & 0.778 & 0.840 & 0.660 & 0.891 & 3.38(2.01) \\
    AutoInt & 0.671 & 0.825 & 0.563 & 0.893 & 0.769 & 0.836 & 0.676 & 0.887 & 4.75(1.25) \\
    \midrule
    TabTrans & 0.653 & 0.732 & 0.584 & 0.856 & 0.784 & 0.792 & 0.674 & 0.828 & 7.62(3.78) \\
    FT-Trans & 0.662 & 0.824 & 0.626 & 0.892 & 0.768 & 0.840 & 0.645 & 0.889 & 4.81(2.59) \\
    \midrule
    VIME  & 0.621 & 0.697 & 0.571 & 0.892 & 0.769 & 0.803 & 0.683 & 0.881 & 7.00(3.01) \\
    SCARF & 0.651 & 0.753 & 0.556 & 0.891 & 0.703 & 0.829 & 0.680 & 0.887 & 6.56(1.32) \\
    \midrule
    \method & \textbf{0.741} & \textbf{0.879} & \textbf{0.665} & \textbf{0.894} & \textbf{0.791} & \textbf{0.841} & \textbf{0.739} & 0.897 & \textbf{1.12(0.35)} \\
    \bottomrule
    \end{tabular}%
  \label{tab:incremental_open}%
\end{table}%

\begin{table}[t]
  \centering
  \caption{Test AUROC results on public datasets under \textbf{transfer learning} across tables. }
    \resizebox{\textwidth}{!}{%
    \begin{tabular}{lcccccccccccccccc|l}
    \toprule
    Methods & \multicolumn{2}{c}{CG} & \multicolumn{2}{c}{CA} & \multicolumn{2}{c}{DS} & \multicolumn{2}{c}{AD} & \multicolumn{2}{c}{CB} & \multicolumn{2}{c}{BL} & \multicolumn{2}{c}{IO} & \multicolumn{2}{c|}{IC} & Rank(Std) \\
          & set1  & set2  & set1  & set2  & set1  & set2  & set1  & set2  & set1  & set2  & set1  & set2  & set1  & set2  & set1  & set2  &  \\
    \midrule
    LR    & 0.69  & 0.69  & 0.81  & 0.82  & 0.47  & 0.56  & 0.81  & 0.81  & 0.68  & 0.78  & 0.77  & 0.82  & 0.71  & 0.81  & 0.81  & 0.84  & 8.57(2.83) \\
    XGB & 0.72  & 0.71  & 0.85  & 0.87  & 0.46  & 0.63  & 0.88  & 0.89  & \textbf{0.80} & 0.81  & 0.76  & 0.82  & 0.65  & 0.74  & \textbf{0.92} & \textbf{0.91} & 5.57(3.62) \\
    MLP   & 0.67  & 0.70  & 0.82  & 0.86  & 0.53  & \textbf{0.67} & 0.89  & 0.90  & 0.73  & \textbf{0.82} & 0.79  & 0.83  & 0.70  & 0.78  & 0.90  & 0.90  & 5.00(2.86) \\
    SNN   & 0.66  & 0.63  & 0.85  & 0.83  & 0.54  & 0.42  & 0.87  & 0.88  & 0.57  & 0.54  & 0.77  & 0.82  & 0.69  & 0.78  & 0.87  & 0.88  & 8.67(2.92) \\
    TabNet & 0.60  & 0.47  & 0.66  & 0.68  & 0.54  & 0.53  & 0.87  & 0.88  & 0.58  & 0.62  & 0.75  & 0.83  & 0.62  & 0.71  & 0.88  & 0.89  & 9.87(2.92) \\
    \midrule
    DCN   & 0.69  & 0.70  & 0.83  & 0.85  & 0.51  & 0.58  & 0.88  & 0.74  & 0.79  & 0.78  & 0.79  & 0.76  & 0.70  & 0.71  & 0.91  & 0.90  & 6.67(2.94) \\
    AutoInt & 0.70  & 0.70  & 0.82  & 0.86  & 0.49  & 0.55  & 0.88  & 0.74  & 0.77  & 0.79  & \textbf{0.79} & 0.76  & 0.71  & 0.72  & 0.91  & 0.90  & 6.03(3.23) \\
    \midrule
    TabTrans & 0.72  & 0.72  & 0.84  & 0.86  & 0.54  & 0.57  & 0.88  & 0.90  & 0.73  & 0.79  & 0.78  & 0.81  & 0.67  & 0.71  & 0.88  & 0.88  & 6.03(2.93) \\
    FT-Trans & 0.72  & 0.71  & 0.83  & 0.85  & 0.53  & 0.64  & \textbf{0.89} & 0.90  & 0.76  & 0.79  & 0.78  & 0.84  & 0.68  & 0.78  & 0.91  & 0.91  & 4.97(1.95) \\
    \midrule
    VIME  & 0.59  & 0.70  & 0.79  & 0.76  & 0.45  & 0.53  & 0.88  & 0.90  & 0.65  & 0.81  & 0.58  & 0.83  & 0.67  & 0.70  & 0.90  & 0.90  & 8.83(3.24) \\
    SCARF & 0.69  & 0.72  & 0.82  & 0.85  & \textbf{0.55} & 0.64  & 0.88  & 0.89  & 0.77  & 0.73  & 0.78  & 0.83  & 0.71  & 0.75  & 0.90  & 0.89  & 5.47(2.42) \\
    \midrule
    \method & \textbf{0.74} & \textbf{0.76} & \textbf{0.87} & \textbf{0.89} & 0.55  & 0.66  & 0.88  & \textbf{0.90} & 0.80  & 0.80  & 0.79  & \textbf{0.84} & \textbf{0.73} & \textbf{0.82} & 0.91  & 0.91  & \textbf{2.33(2.10)} \\
    \bottomrule
    \end{tabular}%
    }
  \label{tab:transfer_open}%
\end{table}%

\begin{table}[t]
  \centering
  \caption{Test AUROC results on public datasets under \textbf{zero-shot learning} setting.}
    \begin{tabular}{lllllllll}
    \toprule
    \method & CG    & CA    & DS    & AD    & CB    & BL    & IO    & IC \\
    \midrule
    Supervised   & 0.581 & 0.635 & 0.571 & 0.898 & 0.733 & 0.822 & 0.702 & 0.875 \\
    Transfer & 0.719 & 0.758 & 0.561 & 0.900 & 0.854 & 0.831 & 0.761 & 0.880 \\
    Zero-shot & 0.685 & 0.721 & 0.538 & 0.892 & 0.710 & 0.804 & 0.742 & 0.874 \\
    \bottomrule
    \end{tabular}%
  \label{tab:zeroshot_open}%
\end{table}%

\begin{table}[t]
  \centering
  \caption{Test AUROC on public datasets under the \textbf{across-table pretraining plus finetuning} setting. \textit{Supervised}: baseline supervised model; \textit{Transfer}: vanilla supervised transfer learning. Red shows the one worse than the baseline \textit{Supervised}.}
    \begin{tabular}{lcccccccc}
    \toprule
    \method & CG    & CA    & DS    & AD    & CB    & BL    & IO    & IC \\
    \midrule
    Supervised    & 0.763 & 0.858 & 0.630 & 0.907 & 0.841 & 0.844 & 0.821 & 0.919 \\
    Transfer   & \textbf{0.786} & \textbf{0.861} & \textbf{0.653} & 0.907 & \textcolor[rgb]{ 1,  0,  0}{0.819} & \textcolor[rgb]{ 1,  0,  0}{0.843} & \textcolor[rgb]{ 1,  0,  0}{0.813} & \textcolor[rgb]{ 1,  0,  0}{0.918} \\
    \midrule
    Self-\clmethod & 0.777 & \textcolor[rgb]{ 1,  0,  0}{0.837} & \textcolor[rgb]{ 1,  0,  0}{0.626} & 0.907 & \textcolor[rgb]{ 1,  0,  0}{0.819} & \textcolor[rgb]{ 1,  0,  0}{0.843} & \textbf{0.823} & \textcolor[rgb]{ 1,  0,  0}{0.919} \\
    \midrule
    \clmethod & 0.776 & 0.858 & 0.637 & \textbf{0.907} & \textbf{0.862} & \textbf{0.844} & \textcolor[rgb]{ 1,  0,  0}{0.819} & \textbf{0.919} \\
    \bottomrule
    \end{tabular}%
  \label{tab:pretrain_open}%
\end{table}%

\begin{figure}[t]
\centering
\begin{minipage}[t]{0.49\textwidth}
\centering
\includegraphics[width=\textwidth]{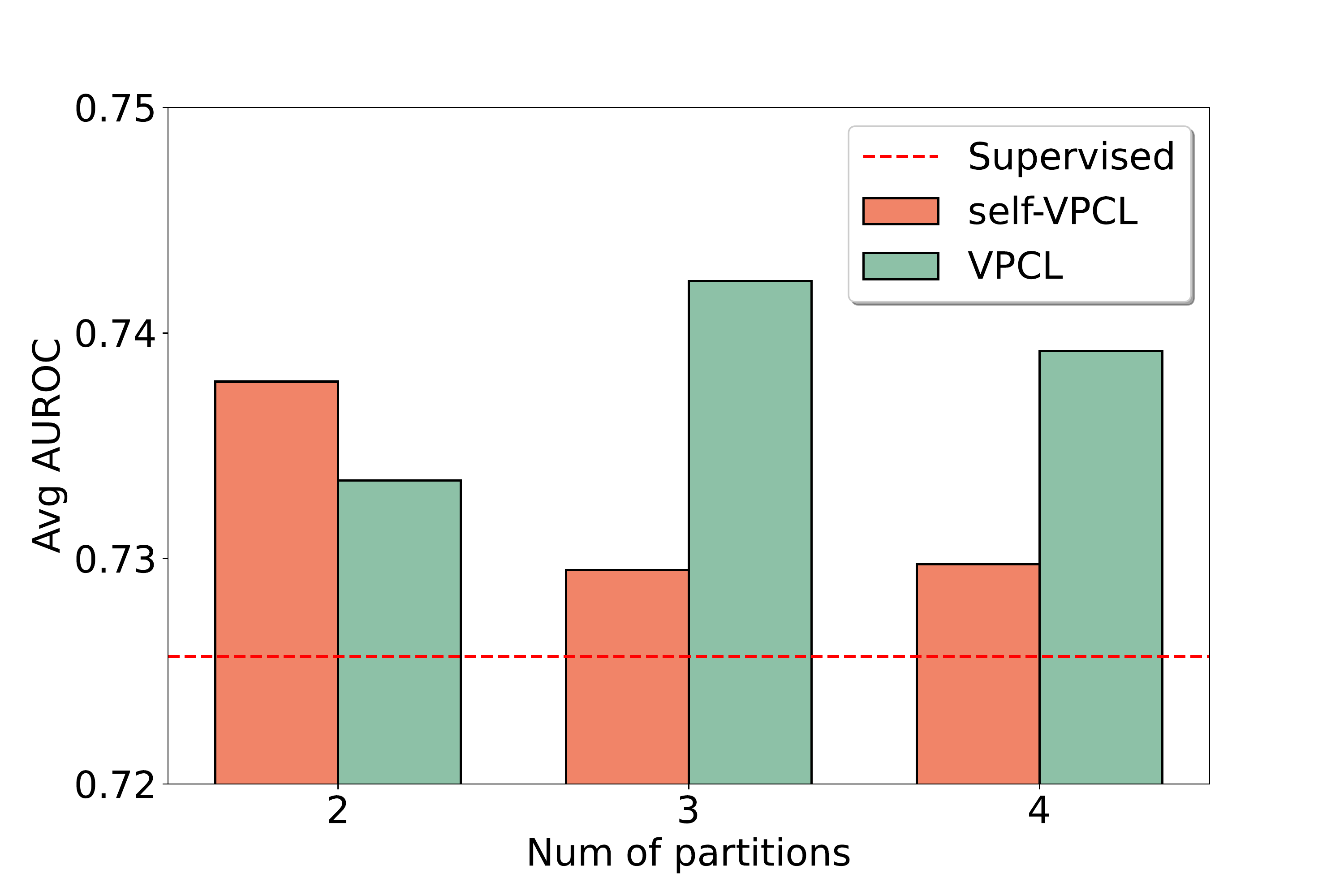}
\caption{Analysis of the number of partitions for \clmethod and self-\clmethod on the \textbf{clinical trial datasets}.  \label{fig:ablation_n_partition_trial}}
\end{minipage}
\hfill
\begin{minipage}[t]{0.49\textwidth}
\centering
\includegraphics[width=\textwidth]{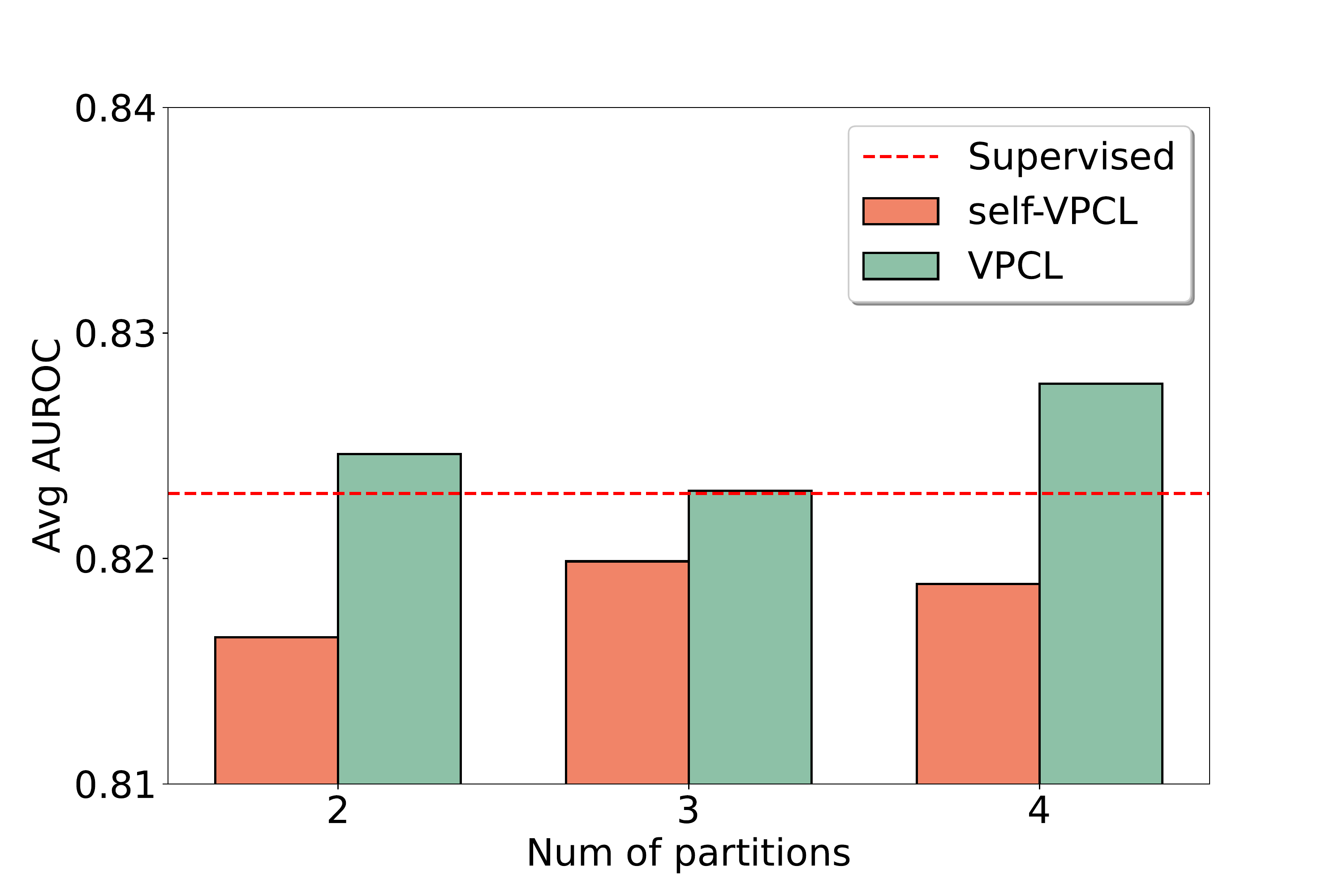}
\caption{Analysis of the number of partitions for \clmethod and self-\clmethod on the \textbf{public datasets} \label{fig:ablation_n_partition_open}}
\end{minipage}
\end{figure}

\begin{table}[t]
  \centering
  \caption{Benchmark dataset links.}
  \resizebox{\textwidth}{!}{%
    \begin{tabular}{ll}
    \toprule
    Dataset & URL \\
    \midrule
    credit-g & \url{https://www.openml.org/search?type=data\&status=active\&id=31} \\
    credit-approval & \url{https://archive.ics.uci.edu/ml/datasets/credit+approval} \\
    dress-sales & \url{https://www.openml.org/search?type=data\&status=active\&id=23381} \\
    adult & \url{https://www.openml.org/search?type=data\&status=active\&id=1590} \\
    cylinder-bands & \url{https://www.openml.org/search?type=data\&status=active\&id=6332} \\
    blastchar & \url{https://www.kaggle.com/datasets/blastchar/telco-customer-churn} \\
    insurance-co & \url{https://archive.ics.uci.edu/ml/datasets/Insurance+Company+Benchmark+\%28COIL+2000\%29} \\
    1995-income & \url{https://www.kaggle.com/datasets/lodetomasi1995/income-classification} \\
    \bottomrule
    \end{tabular}%
    }
  \label{tab:data_url}%
\end{table}%

\begin{figure}[t]
\centering
\begin{subfigure}[b]{0.49\textwidth}
\includegraphics[width=\textwidth]{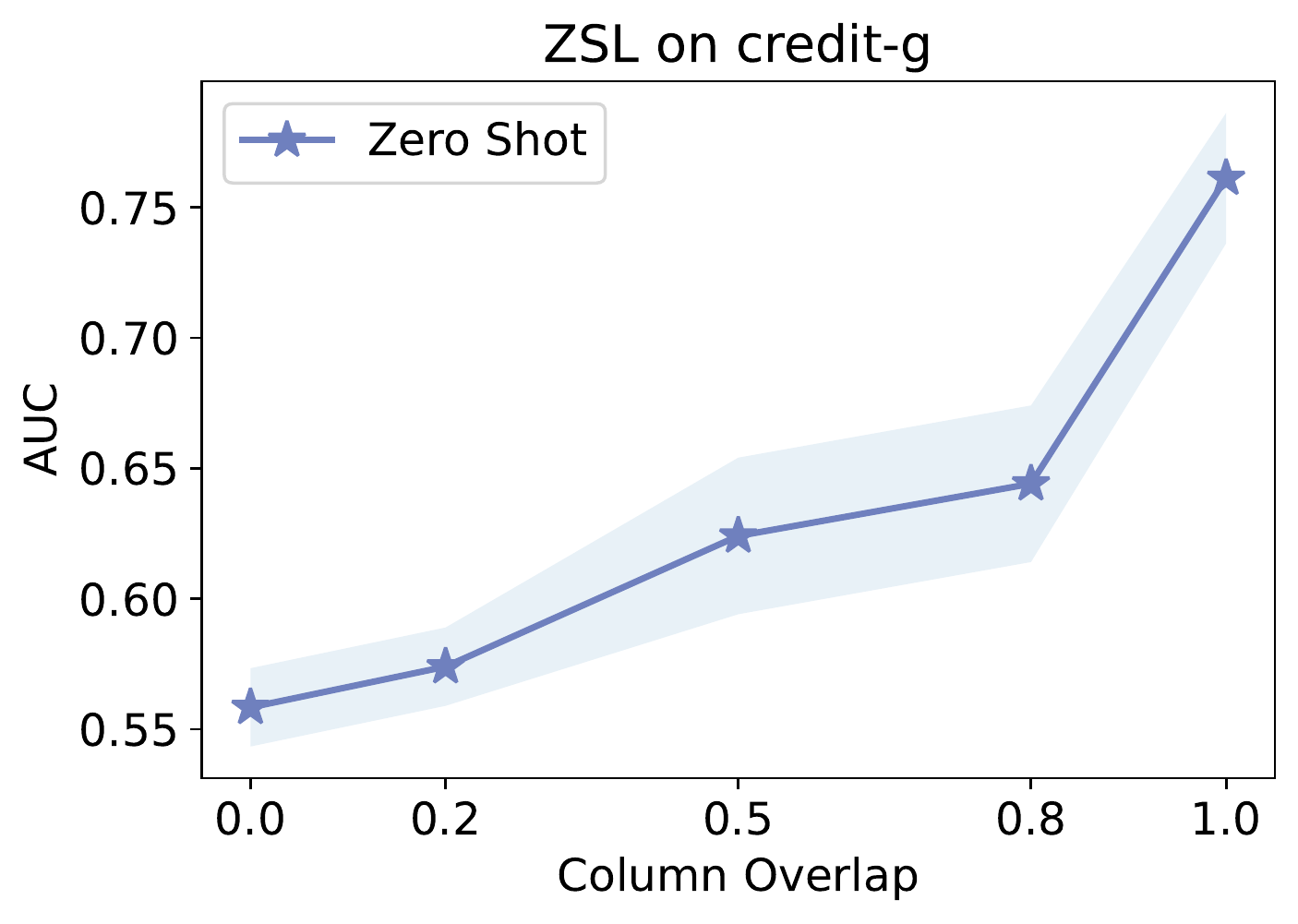}
\end{subfigure}
\hfill
\begin{subfigure}[b]{0.49\textwidth}
\includegraphics[width=\textwidth]{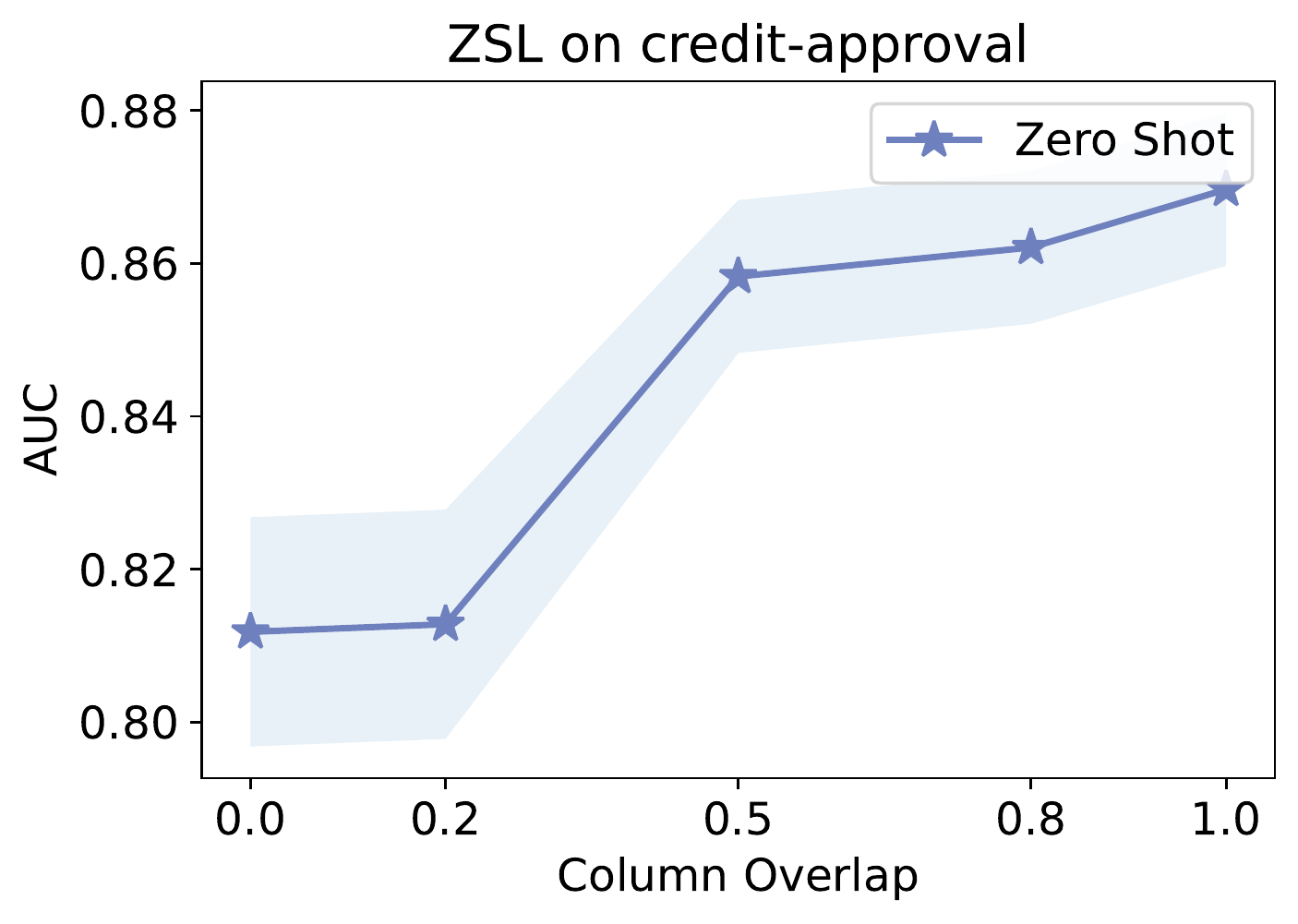}
\end{subfigure}
\begin{subfigure}[b]{0.49\textwidth}
\includegraphics[width=\textwidth]{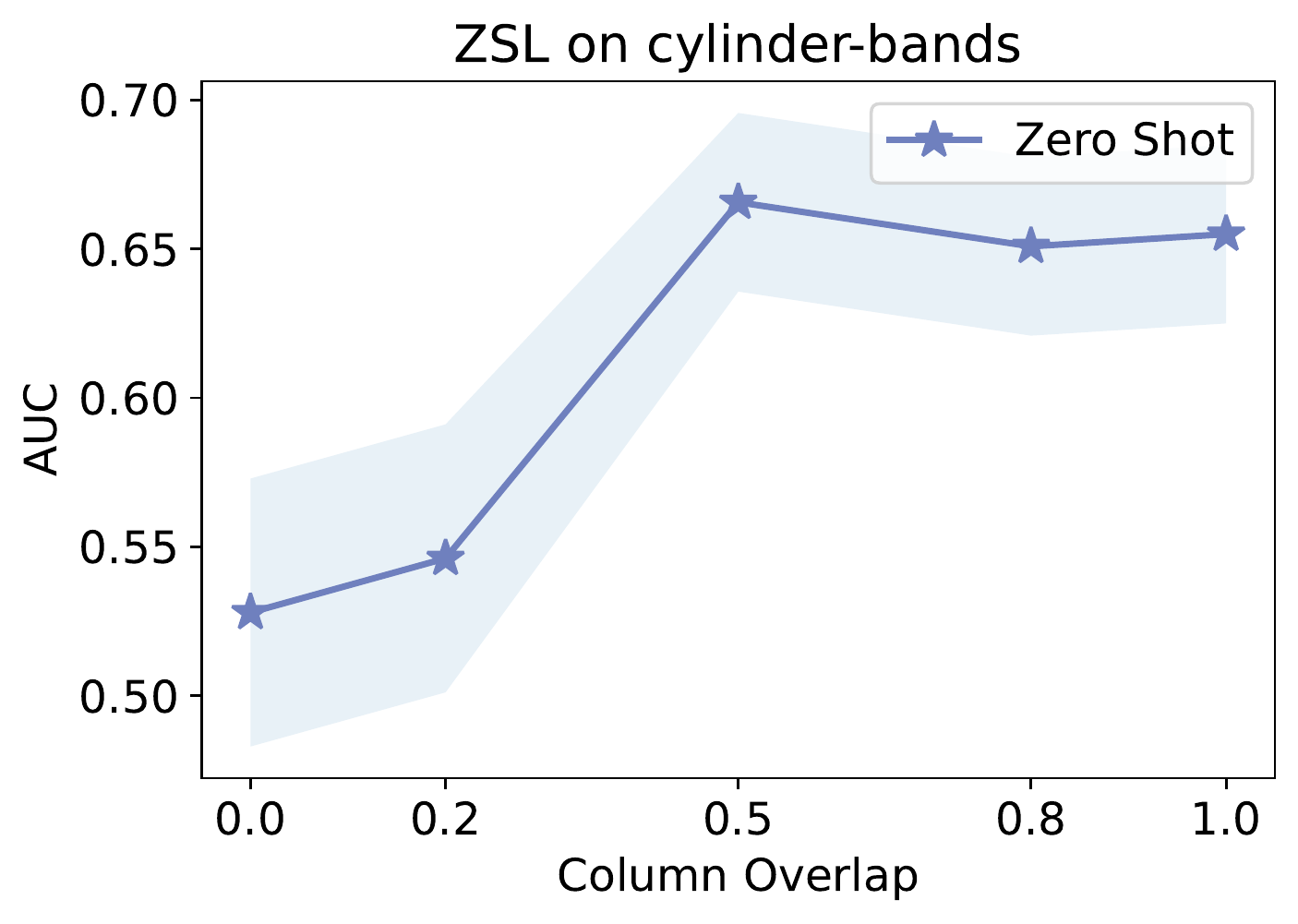}
\end{subfigure}
\begin{subfigure}[b]{0.49\textwidth}
\includegraphics[width=\textwidth]{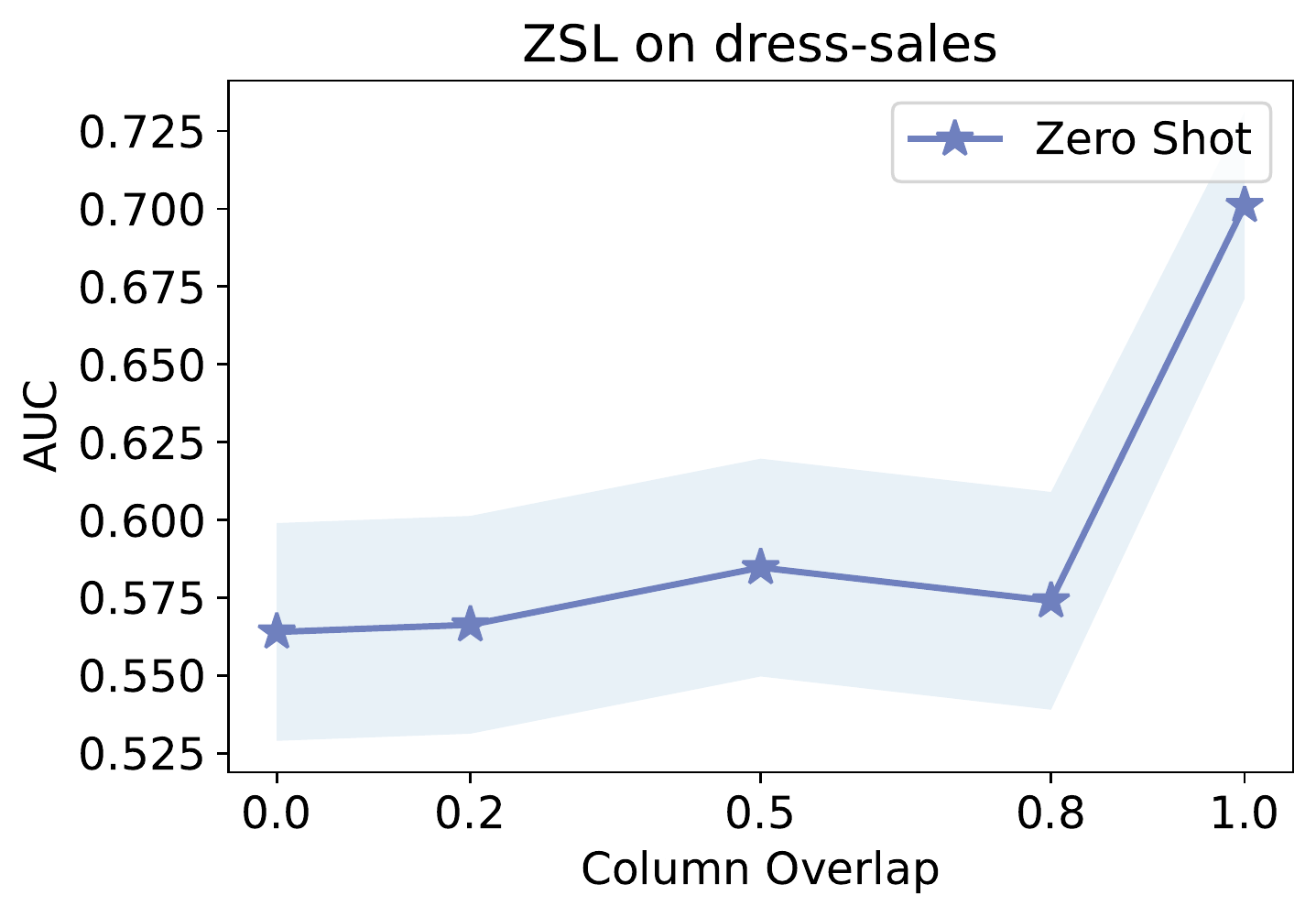}
\end{subfigure}
\caption{Evaluate how the overlap ratio of two tables' columns influences zero-shot learning (ZSL) performance of \method, on public data CG, CA, CB, and DS. $x$-axis: the ratio of test table columns exist in the training table (0: no test table column appears in training table; 1: all test table columns in training table); $y$-axis: the test AUC when making ZSL. \label{fig:zsl_sensitivity}}
\end{figure}

\section{Preprocessing of clinical trial datasets}\label{appx:trial_data}
The raw real-world de-identified patient-level clinical records are obtained from \href{https://data.projectdatasphere.org/projectdatasphere/html/home}{Project Data Sphere}. All clinical trial data used in this paper are available under registration for the data platform. For each trial, we manage to extract patient's baseline information as the features, including demographic information, medical history, medication history, lab test, vital signs, adverse events, etc. We draw the target labels from the survival analysis section where censoring is considered as "alive" and other events are tagged "mortality" so as to transform the datasets into binary prediction tasks.

\section{Establishment of subsets} \label{appx:subset}
For experiments in \S \ref{sec:fil}, \S \ref{sec:tl}, \S \ref{sec:zsl}, we create subsets randomly with a fixed seed, respectively. That is, subsets vary across these experiments.

\begin{itemize}
 \item Feature incremental learning. The columns are splitted into three distinct parts ${v_1,v_2,v_3}$. Set1 contains $v_1$, set2 contains $v_1,v_2$, and set3 has $v_1,v_2,v_3$. Three sets have the equal number of samples.
 \item Transfer learning. The columns are splitted into two parts $v_1,v_2$ where $v_1$ and $v_2$ have 50\% of elements overlapped. Two sets have the equal number of samples.
 \item Zeroshot learning. The columns are splitted into three distinct parts ${v_1,v_2,v_3}$. Set1 contains $v_1$, set2 contains $v_2$, set3 contains $v_3$. Three sets have the equal number of samples.\end{itemize}

\end{document}